\documentclass{article} 
\usepackage[T1]{fontenc}
\usepackage{iclr2027_conference,times}
\iclrfinalcopy
\AddToHook{cmd/maketitle/after}{\fancyhead{}\renewcommand{\headrulewidth}{0pt}}

\usepackage{amsmath,amsfonts,bm}

\def\eqref#1{equation~\ref{#1}}

\def\1{\bm{1}}

\DeclareMathAlphabet{\mathsfit}{\encodingdefault}{\sfdefault}{m}{sl}
\SetMathAlphabet{\mathsfit}{bold}{\encodingdefault}{\sfdefault}{bx}{n}

\usepackage{hyperref}
\hypersetup{hidelinks}
\usepackage{xurl}
\usepackage{booktabs}
\usepackage{tabularx}
\usepackage{graphicx}
\usepackage{placeins}
\usepackage{amssymb}
\usepackage{xcolor}
\usepackage{colortbl}
\definecolor{rankgold}{HTML}{FBF5E5}
\definecolor{ranksilver}{HTML}{F0F3F6}
\definecolor{rankbronze}{HTML}{F8EFE8}

\newcommand{\yesmark}{\textcolor{green!45!black}{$\checkmark$}}
\newcommand{\nomark}{\textcolor{red!65!black}{$\times$}}

\title{LEVI: Benchmarking Visual Identity Fidelity\\
with Verifiable Low-Entropy Constructs}

\author{%
Xiao Ma$^{1}$ \quad Zhonghao Zhan$^{2}$ \quad Yi Tang$^{1}$ \quad Qi Chen$^{1}$\thanks{Corresponding author: \href{mailto:chenqi@molardata.com}{chenqi@molardata.com}.}\\[3pt]
{\normalfont\small $^{1}$ZODA \quad $^{2}$Imperial College London}%
}

\begin{document}

\maketitle

\begin{abstract}
Modern generative models can produce visually plausible images while violating
exact identity constraints. Evaluating such failures requires distinguishing
violations of the requested identity from permissible variation in appearance.
We introduce Low-Entropy Visual Identity (LEVI), which characterizes visual
identities determined by compact facts and explicit rules.
The low-entropy requirement concerns identity uncertainty, while appearance
may remain diverse. We formulate identity fidelity in terms of the probability
that a conditional image distribution assigns to the identities permitted by
the prompt. Visual Identity Spaces (VIS) formalize
this distinction by specifying admissible identity values and which
different realizations express the same identity. Building on this
formulation, we develop Verifiable Visual Grammar (VVG), a declarative
language whose typed primitives and compositional constraints describe
identity-bearing structures. Using this grammar, we build a
benchmark for image generation. It assesses each output against the task's
identity requirements, drawing evidence from the final pixels.
The curated benchmark contains 1,375 tasks spanning 120 base identity
spaces across six domains. Among ten evaluated model configurations,
the highest whole-task pass rate is 58.69\%.
These results reveal a persistent gap between visually plausible images
and faithful realization of their specified identities.
Project page: \url{https://maxiao0234.github.io/LEVI/}.
\end{abstract}

\section{Introduction}

\begin{figure}[t]
  \centering
  \includegraphics[width=\textwidth]{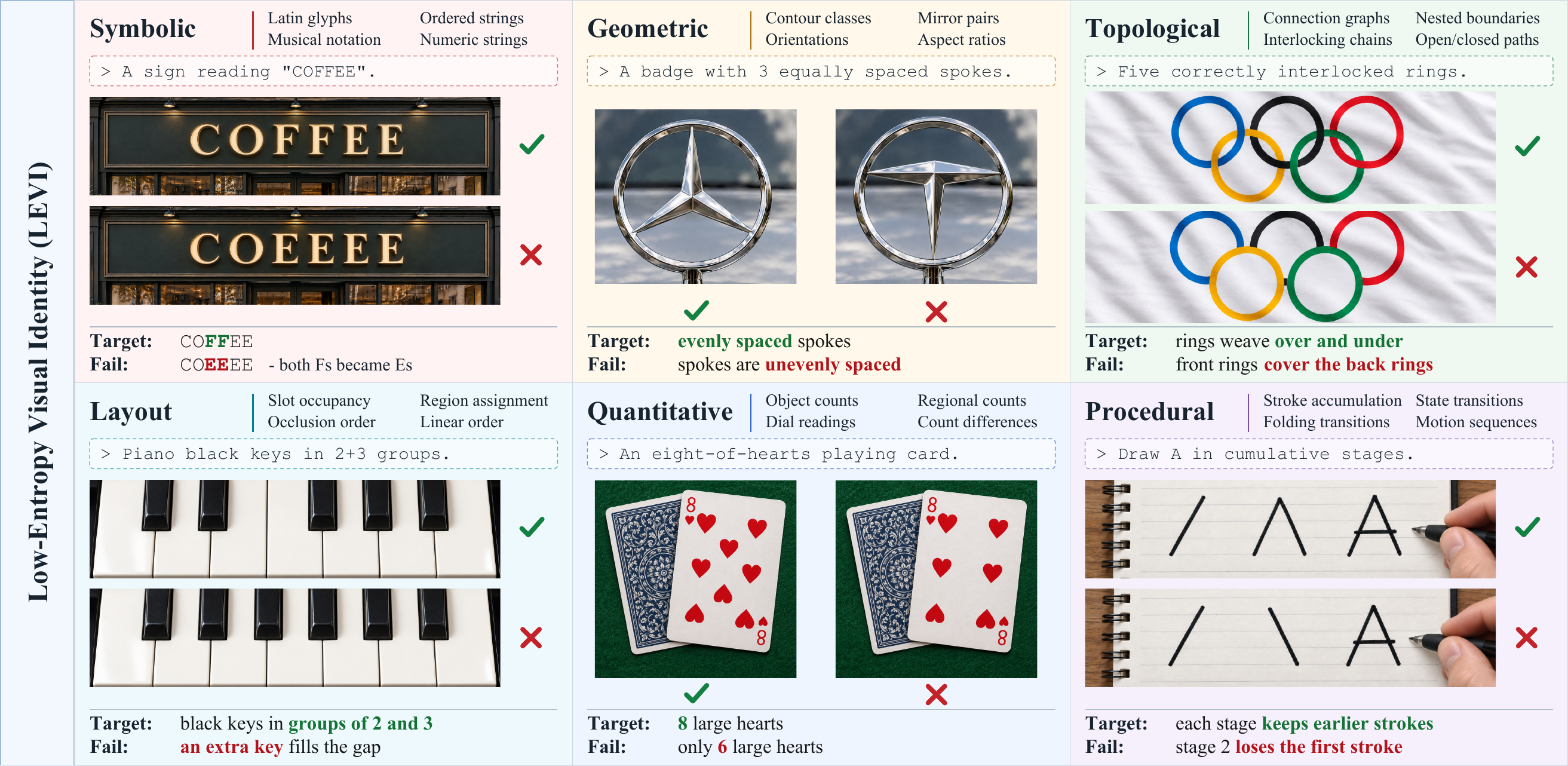}
  \caption{Visual plausibility can coexist with identity errors.
  Each domain pairs matching (\yesmark) and failing (\nomark) examples for
  one prompt. Target/Fail rows identify the mismatch; header entries list
  representative VIS. All examples are conceptual.}
  \label{fig:benchmark-at-a-glance}
\end{figure}

Modern generative models have advanced in visual quality and instruction
following, translating complex text requests into detailed, coherent images.
Their capabilities now extend from natural scene synthesis to visual
creation with more demanding content and layout requirements, such as
integrating text and graphics into posters and infographics
\citep{google2026nanobanana2,seed2026seedream5pro}.
Improvements in overall presentation, however, do not always ensure that
critical details satisfy the request
\citep{zhang2025strict,kamath2025geneval2}.
Figure~\ref{fig:benchmark-at-a-glance} illustrates plausible images whose
identities are altered: a sign retains a realistic appearance while its
character sequence changes. Similar discrepancies affect objects whose
identities depend on geometry or relations among their elements. These
details may occupy a small part of the image yet change what the object
expresses, requiring assessment beyond overall visual quality.

These failures involve a distinction between identity certainty and
appearance diversity. 
For the text in Figure~\ref{fig:benchmark-at-a-glance}, the font or material
may vary while the specified character sequence remains fixed. Global metrics such as FID and CLIPScore \citep{heusel2017gans,hessel2021clipscore} and reference-based similarity do not separate a wrong identity from a legitimate change in appearance, and recent benchmarks with executable checks \citep{zhang2025strict,liu2026mathgen,zhao2026genpuzzle} do not define when
different-looking outputs express the same target (Table~\ref{tab:benchmark-comparison}).
We model visual identity explicitly and use the conditional
entropy chain rule to distinguish these two sources of uncertainty.
When a condition fully specifies the identity,
the desired identity uncertainty is zero, although appearance can remain
diverse. Since a model that always draws the same wrong word also has zero identity entropy, we measure identity fidelity: the probability of realizing an identity permitted by the prompt, which gives equal credit to every valid appearance. Evaluating this property requires specifying which facts determine identity and which changes preserve it.

We introduce Low-Entropy Visual Identity (LEVI) , a benchmark for image generation
in which identity is tightly constrained while appearance remains diverse.
To specify what must be preserved, we formalize Visual Identity Spaces
(VIS), which describe admissible identity values and define when different
images express the same identity. This distinction allows us to assess
whether an image preserves a specified identity while accepting permissible
variation in appearance. A VIS may describe a
standalone identity, while compatible spaces can be composed under shared
constraints to express identities that depend on several components.
Defined transformations and projections relate identities within and
across spaces.

To turn these spaces into benchmark tasks, we develop Verifiable Visual
Grammar (VVG), a declarative language for describing identity-bearing
structures. Its typed components and rules specify how elements can be
combined and which relations must hold, giving each description an
identity under VIS semantics. Using VVG, we construct a benchmark spanning
six identity domains: Symbolic, Geometric, Topological, Layout,
Quantitative, and Procedural, illustrated in
Figure~\ref{fig:benchmark-at-a-glance}. Tasks test both individual and
combined identity requirements. Evaluation reads evidence from the final
image and checks whether it satisfies the target specified at construction,
linking task definition and scoring through the same identity semantics.
Table~\ref{tab:benchmark-comparison} compares LEVI with existing
image-generation benchmarks in terms of visual targets, rule requirements,
verification, and identity modeling. At the identity level, we examine
whether a benchmark supports constructing verifiable visual structures
with explicit semantics, expressing standalone and composed identity
requirements, and treating visually different realizations of the same
identity as equivalent during evaluation.

Our contributions are as follows:
\begin{itemize}
  \item We introduce LEVI and formalize visual identity fidelity. By modeling
  visual identity, we use conditional entropy to distinguish identity
  uncertainty from appearance diversity. VIS specifies admissible identities
  and their equivalent realizations, allowing fidelity to be defined as the
  probability of realizing the specified identity.
  \item We develop VVG to turn identity semantics into constructible and
  verifiable visual requirements. Typed components, relations, and
  compositional constraints express identity-bearing structures, supporting
  task construction from standalone and composed identity requirements and
  verification of generated images under the same semantics.
  \item We construct the LEVI benchmark with 120 base identity spaces and
  1,375 curated tasks across six domains. Among ten evaluated model
  configurations, the highest whole-task pass rate is 58.69\%.
  Identity-space analysis reveals differences obscured by overall rankings,
  while composition counts alone do not explain difficulty.
\end{itemize}

\begin{table}[t]
  \caption{Comparison with recent image-generation benchmarks.
  \yesmark\ denotes explicit support in the primary protocol and
  \nomark\ denotes no explicit support. LEVI denotes the proposed design.}
  \label{tab:benchmark-comparison}
  \centering
  \footnotesize
  \setlength{\tabcolsep}{0.7pt}
  \renewcommand{\arraystretch}{0.98}
  \begin{tabularx}{\textwidth}{@{}>{\raggedright\arraybackslash}p{0.22\textwidth}*{6}{>{\centering\arraybackslash}X}@{}}
    \toprule
    Benchmark &
    {\scriptsize\shortstack{Structured\\domains}} &
    {\scriptsize\shortstack{Rule-governed\\targets}} &
    {\scriptsize\shortstack{Executable\\checks}} &
    {\scriptsize\shortstack{Verifiable\\construction}} &
    {\scriptsize\shortstack{Identity-space\\expressivity}} &
    {\scriptsize\shortstack{Identity\\equivalence}} \\
    \midrule
    STRICT &
    \nomark & \nomark & \yesmark & \nomark & \nomark & \nomark \\
    MMMG &
    \yesmark & \nomark & \yesmark & \nomark & \nomark & \nomark \\
    StructBench &
    \yesmark & \nomark & \nomark & \nomark & \nomark & \nomark \\
    SpatialGenEval &
    \nomark & \nomark & \nomark & \nomark & \nomark & \nomark \\
    MathGen &
    \nomark & \yesmark & \yesmark & \nomark & \nomark & \nomark \\
    GraphicDesignBench &
    \yesmark & \nomark & \yesmark & \nomark & \nomark & \nomark \\
    RIGBench &
    \yesmark & \yesmark & \nomark & \nomark & \nomark & \nomark \\
    GenPuzzle &
    \yesmark & \yesmark & \yesmark & \nomark & \nomark & \nomark \\
    TRACEBench &
    \nomark & \yesmark & \nomark & \yesmark & \nomark & \nomark \\
    \textbf{LEVI (ours)} &
    \yesmark & \yesmark & \yesmark & \yesmark & \yesmark & \yesmark \\
    \bottomrule
  \end{tabularx}
\end{table}

\section{Related Work}

\noindent\textbf{Image-generation evaluation.}
Image-generation benchmarks increasingly assess whether an image satisfies
specific requirements beyond overall visual quality. Global measures such
as FID and CLIPScore summarize distributional similarity or image--text
alignment \citep{heusel2017gans,hessel2021clipscore}, whereas TIFA, GenEval,
and T2I-CompBench++ examine objects, attributes, counts, and relations
\citep{hu2023tifa,ghosh2023geneval,huang2023t2icompbench}. GenEval~2 and
FineGRAIN further investigate fine-grained assessment and its agreement with
human judgment \citep{kamath2025geneval2,hayes2025finegrain}. Knowledge and
spatial benchmarks extend the target to factual dependencies and spatial
relations \citep{huang2025t2ifactualbench,luo2025mmmg,li2026spatialgeneval}.
Reasoning-oriented evaluations require deriving the visual content from
implicit instructions or contextual rules
\citep{sun2026t2ireasonbench,liu2026rigbench,zhao2026genpuzzle}.
These developments establish the importance of precise visual requirements.
LEVI focuses on their identity semantics: which facts determine the target
and which changes preserve it.

\noindent\textbf{Visual identity and target specification.}
Defining a visual target requires separating identity from its appearance.
Reference-conditioned benchmarks assess whether a generated subject or
product retains the properties of its reference
\citep{peng2025dreambenchplus,singhania2025beyondpixels,
khanna2026productconsistency,zheng2026copycat}. MMIG-Bench combines identity,
semantic, and aesthetic assessment \citep{hua2025mmigbench}.
For text, graphics, and mathematical images, target properties can instead
be specified through strings, structural descriptions, and task-specific
constraints \citep{zhang2025strict,zou2024vgbench,nishina2024svgeditbench,
deganutti2026graphicdesignbench,zhuo2025structuredvisuals,liu2026mathgen}.
TRACE-Bench makes the link between specification and assessment explicit:
a compositional operator formula drives both multi-reference task
construction and aligned evaluation \citep{wang2026tracebench}.
LEVI develops this connection at the level of identity spaces. VIS specifies
admissible identities and equivalent realizations, including standalone
spaces and their compatible compositions; VVG expresses structures under
those semantics. Table~\ref{tab:benchmark-comparison} distinguishes this
identity formulation from task scope, target rules, and evaluation methods.

\noindent\textbf{Verification of generated images.}
Verification must recover the relevant facts from the generated pixels.
ControlNet++ checks re-extracted spatial conditions, while DiffQRCoder
assesses whether a rendered code remains readable
\citep{li2024controlnetpp,liao2024diffqrcoder}. MathGen and ProVisE use
external procedures to assess visual outputs
\citep{liu2026mathgen,wang2026showdonttell}; GenPuzzle combines programmatic
checks for discrete outputs with task-specific visual judging
\citep{zhao2026genpuzzle}. VBVR applies rule-based scorers to generative
visual reasoning \citep{wang2026vbvr}, and SpatialReward connects grounded
observations with spatial relation checks \citep{zhou2026spatialreward}.
The reliability of the observation procedure remains consequential:
MJ-Bench evaluates systematic weaknesses of multimodal judges
\citep{chen2024mjbench}. LEVI separates visual readout from the fixed
identity check. The reader supplies observed evidence, and the checker
applies the same identity semantics used in construction, preserving the
distinction between an identity violation and a measurement failure.

\begin{figure}[!ht]
  \centering
  \includegraphics[width=\textwidth]{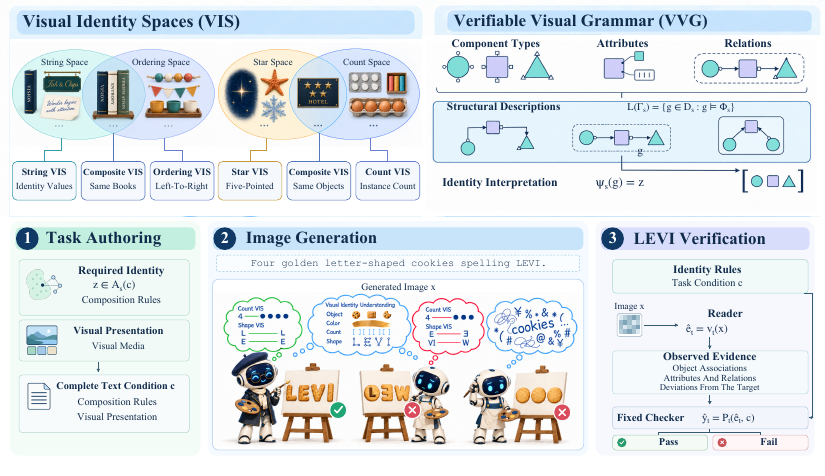}
  \caption{From visual identity to output-based verification.
  Top: VIS groups equivalent realizations and combines compatible identities;
  VVG describes a structure $g$ whose interpretation is an identity $z$.
  Types, attributes, and relations jointly define admissible descriptions.
  Bottom: task authoring supplies a complete
  text condition, generation produces pixels, and verification checks evidence against the shared identity conditions. Robots
  illustrate conceptual distinctions in identity understanding.}
  \label{fig:vis-vvg-overview}
\end{figure}
\FloatBarrier

\section{Method}

We distinguish uncertainty about identity from diversity in its visual
realization. Identity fidelity measures agreement with the target; VIS
organizes admissible values and their equivalence; VVG expresses these
requirements as structures that support construction and evaluation.

\subsection{Low-Entropy Visual Identity}
\label{sec:identity-fidelity}

Image generators model a conditional distribution $q_\theta(X\mid C)$ over
images $X$ given text $C$. The condition constrains visual identity while
allowing variation in its realization. Let $s$ specify the identity rules
and let the deterministic map $\phi_s$ assign each image its expressed
identity, giving $Z_s=\phi_s(X)$. Using discrete representations, with
continuous quantities quantized at a specified resolution, the conditional
entropy chain rule \citep{cover2006} gives
\begin{equation}
  H(X\mid C=c)
  =H(Z_s\mid C=c)+H(X\mid Z_s,C=c).
  \label{eq:entropy-decomposition}
\end{equation}
Under $q_\theta$, the first term measures uncertainty about the expressed
identity, and the second measures variation among realizations of a given
identity. Low-Entropy Visual Identity (LEVI) concerns tightly specified
identities with diverse visual realizations. For an exact target $z^\star$,
the desired generation distribution assigns all probability to its
realizations, giving zero identity entropy while retaining permissible
visual variation.

Entropy measures the concentration of generated identities. Identity
fidelity (IF) measures their agreement with the condition. Writing $A_s(c)$
for the permitted identity values, we define
\begin{equation}
  \mathrm{IF}_\theta(c,s)
  =\Pr_{X\sim q_\theta(\cdot\mid c)}
  \!\left[\phi_s(X)\in A_s(c)\right].
  \label{eq:identity-fidelity}
\end{equation}
For an exact target, $A_s(c)=\{z^\star\}$. This probability aggregates all
permitted realizations, giving equivalent appearances equal credit.

\subsection{Visual Identity Spaces}
\label{sec:vis}

A Visual Identity Space (VIS) specifies which visual differences change
identity. For an ordered sequence of symbols, a painted
$\triangle\,\circ\,\square$ and an engraved version can express the same
identity, while exchanging the circle and square changes it. The declared
rules specify which changes of appearance are allowed. Under identity rules $s$,
let $\mathcal X_s^{\mathrm{adm}}$ contain images expressing admissible
identities. The relation $x\sim_s x'$ groups images that express the same
identity into an equivalence class. These classes form the corresponding
VIS:
\begin{equation}
  \mathcal Z_s=\mathcal X_s^{\mathrm{adm}}/{\sim_s},
  \qquad
  \phi_s(x)=[x]_{\sim_s},\quad
  x\in\mathcal X_s^{\mathrm{adm}}.
  \label{eq:vis-quotient}
\end{equation}
Each point $z\in\mathcal Z_s$ represents one identity, and $\phi_s$
maps its equivalent image realizations to that point. Variation within a
class preserves identity; movement between classes changes it. Images outside
the admissible domain receive the value $\bot$, so
Equation~\ref{eq:identity-fidelity} remains defined for every output.

An identity space can stand alone or combine with other spaces. When
component values $z_1,\ldots,z_m$ completely determine a composite identity,
its space can be written as
\begin{equation}
  \mathcal Z_s =
  \left\{(z_1,\ldots,z_m)\in
  \prod_{j=1}^{m}\mathcal Z_{s_j}:
  \chi_s(z_1,\ldots,z_m)=1\right\},
  \label{eq:vis-composition}
\end{equation}
where $\chi_s=1$ selects combinations that can be jointly realized. This
representation includes component identities and every relation that
determines their joint identity. Components can belong
to the same or different domains; compatibility constraints allow them to
depend on one another. The spaces may be discrete, continuous, or mixed.

Figure~\ref{fig:vis-vvg-overview} gives two concrete combinations. The
book spines jointly specify the strings \texttt{VISION}, \texttt{ENTROPY},
and \texttt{VERIFICATION} and their left-to-right order on the same books.
The hotel sign combines a five-pointed star identity with a count of five
instances of that shape. Composition binds these requirements to the same
objects, retaining both component identities and their relations.
The six identity domains organize these visual facts; each can contain
multiple VIS with its own identity values and equivalence rules.

\subsection{Verifiable Visual Grammar}
\label{sec:tasks}

VIS specifies the meaning of a visual identity. Verifiable Visual Grammar
(VVG) expresses that identity through a declarative structural description.
A grammar $\Gamma_s$ defines component types and interfaces, assigns
attributes to their components, and constrains valid relations and
combinations. In the sequence example, components are symbol occurrences,
each owns a symbol value, and successor relations specify their order.
Constraints require values from the declared alphabet and one complete
linear sequence. A description $g$ fixes these values and relations; the
language $\mathcal L(\Gamma_s)$ contains the descriptions satisfying the
constraints. An interpretation map assigns each description its identity:
\begin{equation}
  \psi_s:\mathcal L(\Gamma_s)\longrightarrow\mathcal Z_s,
  \qquad g\longmapsto z
  \label{eq:vvg-language}
\end{equation}
Here, $g$ specifies a structure and $\psi_s(g)$ gives its meaning in VIS.
Descriptions that map to the same point express the same identity. A target
$z^\star$ can therefore be specified by any valid description satisfying
$\psi_s(g)=z^\star$, allowing construction to vary while preserving the
intended identity. Grammar membership and task acceptance are distinct:
both $\triangle\,\circ\,\square$ and
$\triangle\,\square\,\circ$ are legal sequences, but only the first
satisfies an exact request for that order.

A description is verifiable when each identity-defining attribute and
relation is linked to observable image evidence and an explicit decision
rule. These rules specify what must hold in the generated image, allowing
the constraints used in construction to be checked on its visual
realization. For the sequence, the reader recovers the visible symbols
and their order, and the checker compares this evidence with the requested
sequence under the same identity rules. Figure~\ref{fig:vis-vvg-overview}
connects these semantics to task authoring and output-based verification.

Instantiation assigns component values, composition establishes their
declared relations, and
transformations modify the description consistently with VIS operations.
Each resulting structure therefore has an explicit identity interpretation.

Visual media give the resulting structures concrete form. Their appearance
can vary within the equivalence class of the specified identity. The
required identity and composition rules are expressed in the complete text
condition $c$, together with the visual presentation. This condition is
passed to image generation, while the
structural description supplies the target for subsequent verification.

\subsection{Identity-Grounded Benchmarking}
\label{sec:benchmark}

The benchmark evaluates identity fidelity by connecting the target
specified during construction to the content of the generated image.
For an exact identity, $\psi_s(g)=z^\star$ specifies that target and
$\phi_s(x)=z^\star$ defines successful realization. The target description
determines the attributes and relations relevant to this comparison, while
the equivalence rules specify the visual variation that evaluation accepts.

Each identity requirement is associated with evidence that can be observed
in the final image and a rule for deciding whether it is satisfied. Readout
records the relevant content together with its object associations, so that
attributes and relations can be assigned to the components they describe.
This evidence preserves the generated structure, including its deviations
from the target. The checker then applies the target requirements to these
observations.

Figure~\ref{fig:vis-vvg-overview} connects this evidence-based assessment to
task construction and image generation. For task $t$, a reader $v_t$
extracts evidence $\widehat e_t$ from image $x$. A fixed checker $P_t$
assesses it under the identity rules and task condition $c$. For resolved
evidence,
\begin{equation}
  \widehat e_t=v_t(x),
  \qquad \widehat y_t=P_t(\widehat e_t,c),
  \quad \widehat y_t\in\{0,1\}.
  \label{eq:benchmark-readout}
\end{equation}
The semantic map $\phi_s$ defines the identity expressed by an image, while
the reader and checker provide an evidence-based estimate of agreement with
the target. For a composite identity, the decision combines requirements on
individual components with those on their joint structure. Successful
realization requires these conditions to hold together. The same evidence
also identifies which component or relation accounts for an identity
violation, connecting the overall judgment to interpretable differences in
the output.

Equivalence rules determine how variation in appearance enters this
assessment. Realizations within a permitted identity class satisfy the same
identity criterion even as their appearance varies. Applying these rules
consistently across construction and checking gives model comparison a
common semantic basis, while retaining the identity-specific requirements
of each task.

\section{Evaluation and Analysis}
\label{sec:evaluation}

We evaluate visual identity fidelity at three levels: overall performance
across models, variation across identity domains, and behavior within
individual and composite VIS. Qualitative comparisons connect these scores
to the visible facts that distinguish faithful realizations from errors.

\subsection{Benchmark and Evaluation Setup}
\label{sec:eval-setup}

\begin{figure}[!ht]
  \caption{Domain coverage and top-five model profiles.
  (a) B-VIS catalogue share (grey) and task involvement (colour).
  Whole-task pass rates (\%) are (b) B-VIS-balanced or (c) task-weighted.
  Radar axes share a 0--100\% scale; domain subsets overlap.}
  \label{fig:benchmark-profile}
  \centering
  \includegraphics[width=\textwidth]{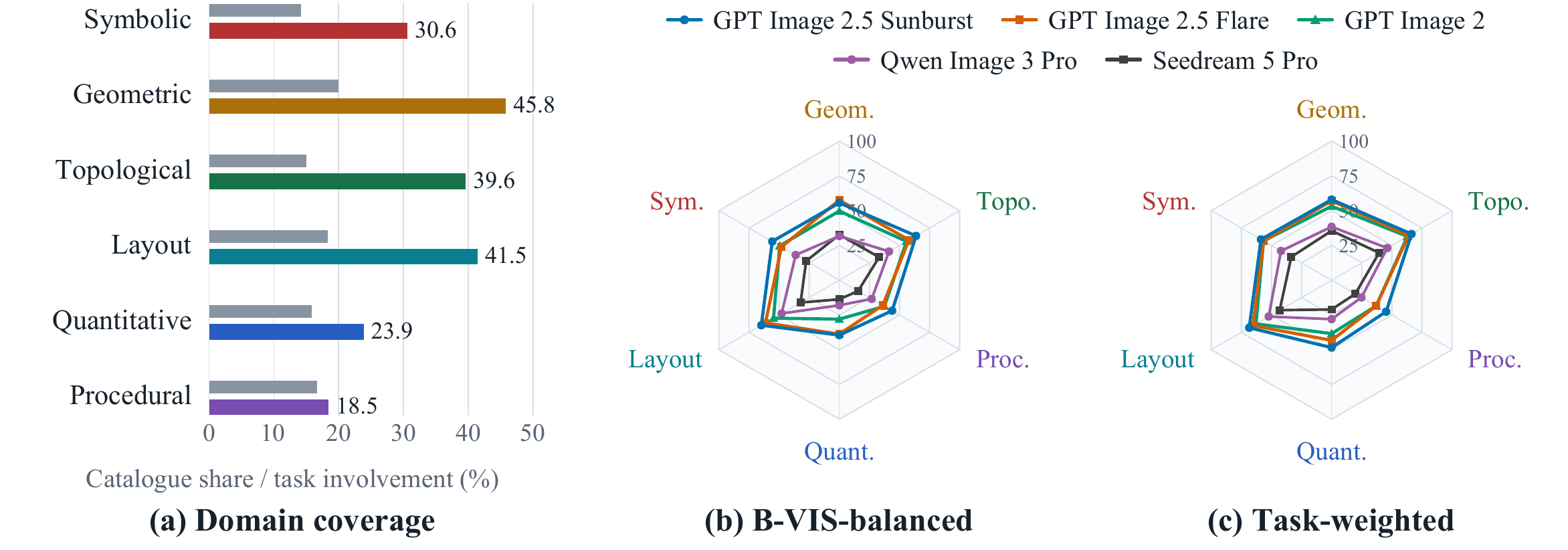}
\end{figure}

The curated benchmark contains 1,375 tasks spanning all 120 base visual
identity spaces (B-VIS), each assigned to one of six domains. Tasks instantiate individual
spaces or combine compatible identities and relations into composite VIS.
A task can therefore contribute to several domain subsets.
Figure~\ref{fig:benchmark-profile} contrasts the catalogue's domain
composition with task involvement in the evaluation set and compares the
models' performance profiles.

The ten configurations in Table~\ref{tab:main-results} each receive the
complete text prompt without reference images or construction records.
Evaluation covers 13,750 model--task cases, with one output per case.

The reported metric is the fraction of retained tasks whose required
conditions all pass automatic assessment. GPT-6-Sol with medium reasoning
effort assesses the generated image against the prompt and its identity
criteria. Missing images and unresolved cases receive zero credit and
remain distinct from observed identity violations.

The challenge set was curated post hoc for redundancy, ambiguity, sampling
balance, and observed ease across models. Reported rates are conditional on
this selection. Task-specific judging instructions include two revisions,
whose equivalence has not been established; their provenance is \mbox{retained}.
Independent human agreement has not been measured. One missing FLUX output
receives zero credit, leaving 13,749 assessed outputs within the full
denominator.

\subsection{Overall Performance}
\label{sec:overall-performance}

Table~\ref{tab:main-results} ranks the ten configurations by overall
whole-task pass rate. GPT Image 2.5 Sunburst leads at 58.69\%, followed by
Flare at 55.71\% and GPT Image 2 at 53.24\%. Qwen Image 3 Pro ranks fourth
at 39.71\%, 13.53 percentage points behind GPT Image 2. Even the leading
configuration fails to receive credit on more than two fifths of tasks.

\begin{table}[!ht]
  \caption{Whole-task pass rates (\%, $\uparrow$) on the post-hoc curated
  challenge set. Medals mark the top three overall;
  \textbf{bold}/\underline{underline} mark the best/second-best per column.}
  \label{tab:main-results}
  \centering
  \small
  \setlength{\tabcolsep}{2.5pt}
  \renewcommand{\arraystretch}{1.28}
\def\LeviPreviewRows{%
\rowcolor{rankgold} \includegraphics[height=2.2ex]{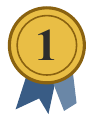} & GPT Image 2.5 Sunburst & \textbf{58.69} & \textbf{58.67} & \textbf{57.94} & \textbf{66.42} & \textbf{68.42} & \textbf{48.48} & \textbf{45.28} \\
\rowcolor{ranksilver} \includegraphics[height=2.2ex]{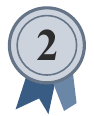} & GPT Image 2.5 Flare & \underline{55.71} & \underline{56.53} & \underline{57.14} & \underline{63.85} & \underline{65.26} & \underline{43.29} & \underline{37.01} \\
\rowcolor{rankbronze} \includegraphics[height=2.2ex]{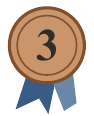} & GPT Image 2 & 53.24 & \underline{56.53} & 53.17 & 62.94 & 62.81 & 38.41 & \underline{37.01} \\
4 & Qwen Image 3 Pro & 39.71 & 42.04 & 38.41 & 46.24 & 52.28 & 28.05 & 24.80 \\
5 & Seedream 5 Pro & 33.67 & 33.49 & 35.56 & 39.45 & 43.16 & 21.04 & 19.69 \\
6 & Nano Banana 2 & 32.44 & 34.44 & 34.76 & 37.25 & 39.30 & 19.82 & 21.65 \\
7 & Nano Banana Pro & 30.91 & 35.87 & 31.75 & 33.76 & 37.89 & 20.43 & 22.44 \\
8 & Grok Image 2 & 28.29 & 34.68 & 27.94 & 33.21 & 34.74 & 19.51 & 18.90 \\
9 & MAI Image 2.5 & 23.49 & 29.69 & 24.92 & 25.32 & 30.18 & 14.63 & 13.39 \\
10 & FLUX.2 Max & 15.27 & 19.71 & 14.76 & 19.08 & 19.82 & 8.54 & 6.69 \\%
}

  \begin{tabularx}{\textwidth}{@{}>{\centering\arraybackslash}p{0.055\textwidth}>{\raggedright\arraybackslash}p{0.26\textwidth}*{7}{>{\centering\arraybackslash}X}@{}}
    \toprule
    Rank & Model & Overall & Sym. & Geom. & Topo. & Layout & Quant. & Proc. \\
    \midrule
    \LeviPreviewRows
    \bottomrule
  \end{tabularx}
\end{table}
\FloatBarrier

Sunburst also leads every task-weighted domain column. Flare places second,
tying GPT Image 2 on Symbolic (56.53\%) and Procedural (37.01\%).

\subsection{Domain-Level Analysis}
\label{sec:domain-analysis}

Figure~\ref{fig:benchmark-profile} presents two complementary views of
LEVI capability. The task-weighted view measures overall task-solving
capability under the evaluated task distribution: success requires all
identity conditions in a task to hold. The B-VIS-balanced view highlights
strengths and weaknesses across LEVI identity categories by giving each
represented B-VIS equal weight within its domain, rather than allowing
frequently sampled identity types to dominate. Both views retain the same
whole-task success criterion.

Reweighting changes the Geometric leader: Flare reaches 57.57\% under
B-VIS balancing, compared with Sunburst's 55.58\%. Sunburst retains the
lead in the other five domains. Thus, the frequency of individual identity
types affects the observed model ordering even within a domain.

The whole-task profiles in Figure~\ref{fig:benchmark-profile}(c) show a common
contraction on Quantitative and Procedural task subsets. Sunburst reaches
68.42\% on Layout and 66.42\% on Topological, compared with 48.48\% on
Quantitative and 45.28\% on Procedural. Every configuration scores lower
on both of the latter subsets than on its Layout subset. These gaps
describe performance on different task populations, rather than independent
measurements of domain-specific reasoning abilities.

The middle of the leaderboard also contains rank reversals. Nano Banana 2
exceeds Nano Banana Pro overall by 1.53 points, yet Pro performs better on
Symbolic (35.87\% versus 34.44\%) and Procedural (22.44\% versus 21.65\%).
Domain membership identifies what a task involves, not which condition
caused its failure. A topology-and-layout task contributes to both subsets
even when only one of its requirements is violated.

\subsection{Identity Spaces and Composition}
\label{sec:vis-analysis}

The mean pass rates across ten models are 35.45\%, 35.46\%, and 38.31\%
for tasks involving one, two, and at least three distinct B-VIS types.
Cross-domain tasks reach 38.24\%, compared with 34.11\% for single-domain
tasks. These are different task populations, not matched component
additions: more identity types do not imply lower success in this set.
Table~\ref{tab:vis-composition} reports the full structure breakdown.
Within domains, Sunburst reaches 60.65\% on tasks involving orientation
but 43.48\% on height order.

\begin{table}[!ht]
  \caption{Whole-task pass rates (\%) by task structure. Share is the
  fraction of tasks in each group; Mean averages all ten models.
  The two groupings partition the same task set separately.}
  \label{tab:vis-composition}
  \centering
  \small
  \setlength{\tabcolsep}{4pt}
  \renewcommand{\arraystretch}{1.18}
  \newcommand{\LeviStructureRows}{%
One B-VIS type & 10.40 & 35.45 & 62.24 & 52.45 & 49.65 \\
Two B-VIS types & 30.62 & 35.46 & 56.06 & 52.97 & 50.59 \\
Three or more B-VIS types & 58.98 & 38.31 & 59.43 & 57.71 & 55.24 \\
\midrule
Single-domain & 26.69 & 34.11 & 55.31 & 50.95 & 50.41 \\
Cross-domain & 73.31 & 38.24 & 59.92 & 57.44 & 54.27 \\
}

  \begin{tabularx}{\textwidth}{@{}Xrrrrr@{}}
    \toprule
    Task structure & Share (\%) & Mean & Sunburst & Flare & GPT Image 2 \\
    \midrule
    \LeviStructureRows
    \bottomrule
  \end{tabularx}
\end{table}

\FloatBarrier
\subsection{Visible Identity Errors}
\label{sec:qualitative}

\begin{figure}[!ht]
  \caption{Identity-level comparisons, one task per row. Vector references
  illustrate the text-only targets; outputs are uncropped.
  Pass/fail denotes whole-task assessment.}
  \label{fig:qualitative-results}
  \centering
  \begingroup
\setlength{\tabcolsep}{2pt}
\renewcommand{\arraystretch}{1.0}
\newcommand{\caseimg}[1]{\includegraphics[width=0.157\textwidth]{figures/identity_cases/#1}}
\newcommand{\casenote}[3]{\textcolor{#1}{\footnotesize\textbf{#2}}}
\begin{tabular}{@{}cccccc@{}}
\footnotesize\textbf{Vector reference} & \shortstack{\footnotesize\textbf{GPT Image 2.5}\\\footnotesize\textbf{Sunburst}} & \footnotesize\textbf{GPT Image 2} & \shortstack{\footnotesize\textbf{Qwen Image}\\\footnotesize\textbf{3 Pro}} & \shortstack{\footnotesize\textbf{Nano Banana}\\\footnotesize\textbf{Pro}} & \footnotesize\textbf{FLUX.2 Max} \\[4pt]
\multicolumn{6}{@{}l@{}}{\scriptsize (a) Two trays: 12 five-point stars in 3-by-4 rows; 10 rings in 2-by-5 rows.} \\[2pt]
\caseimg{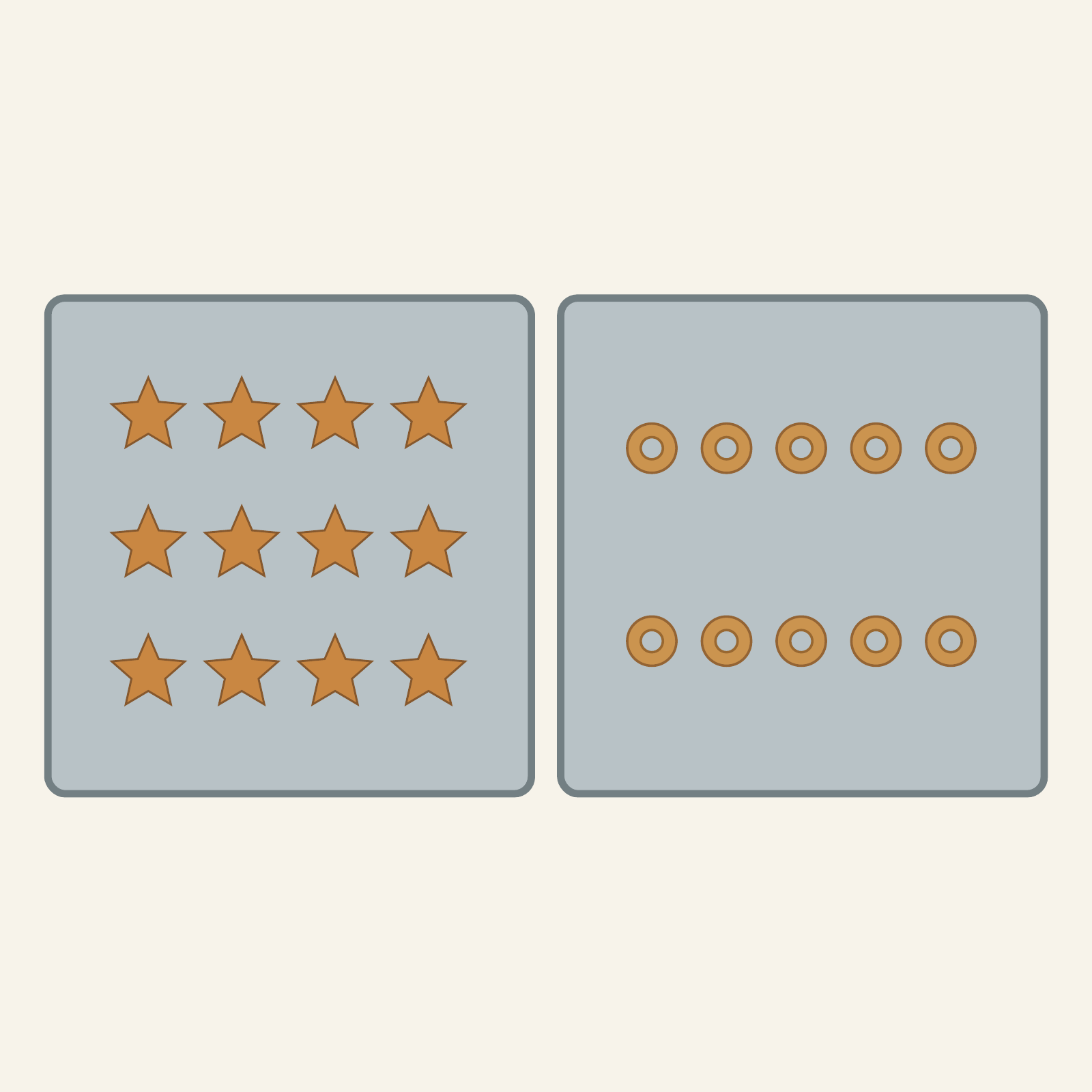} & \caseimg{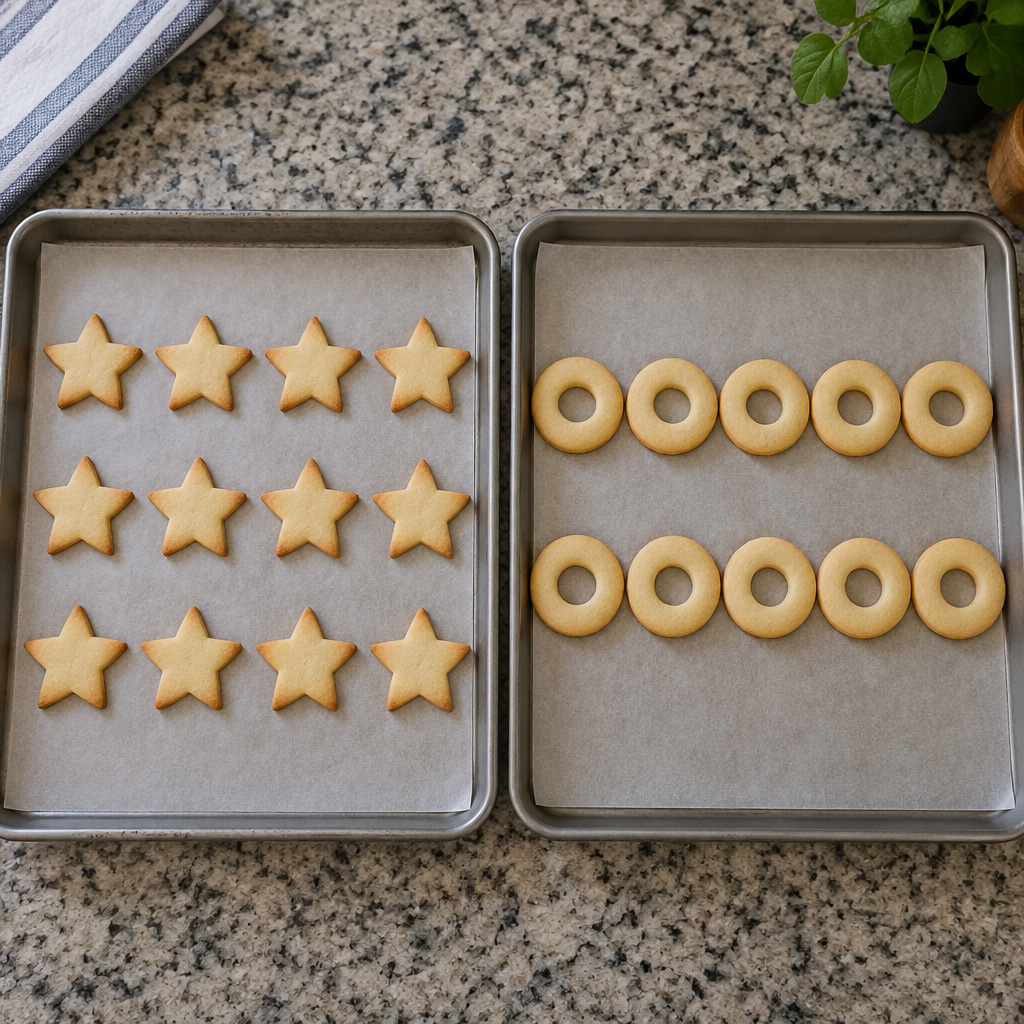} & \caseimg{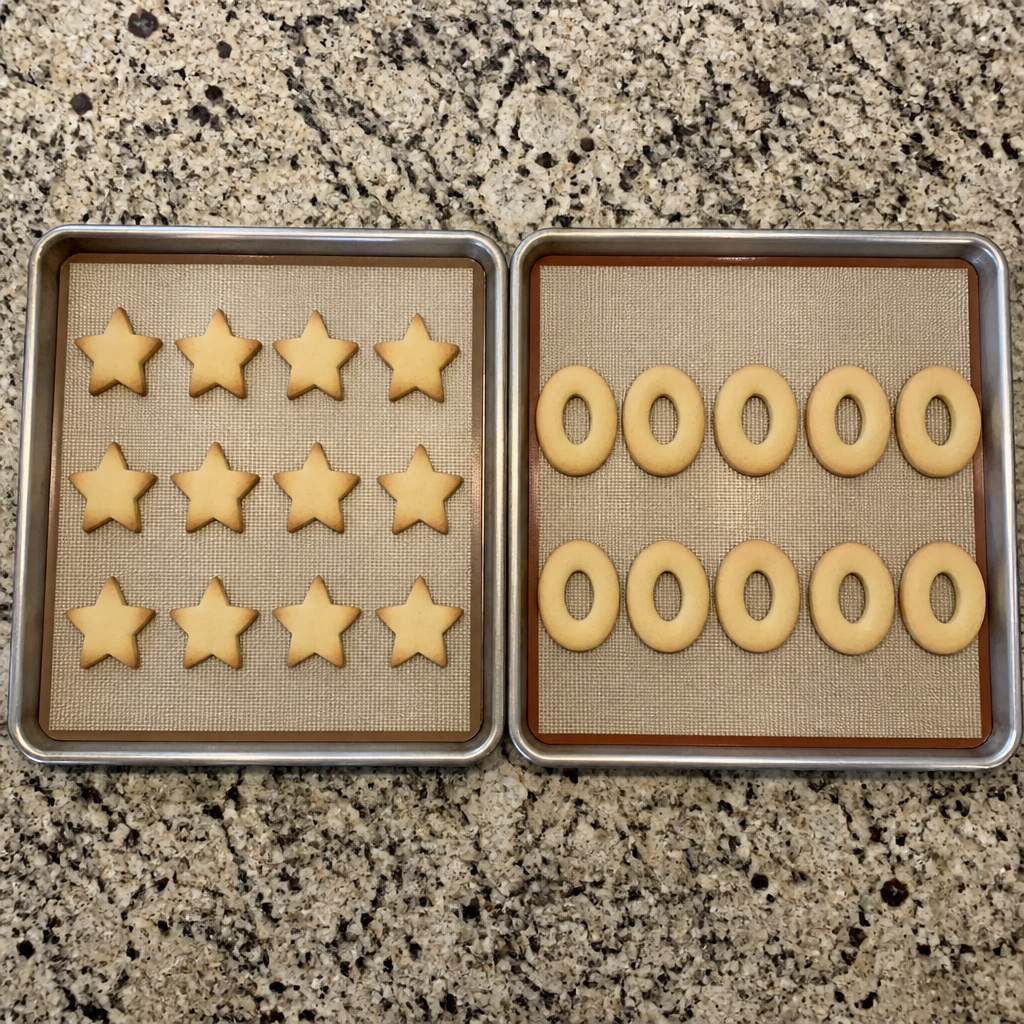} & \caseimg{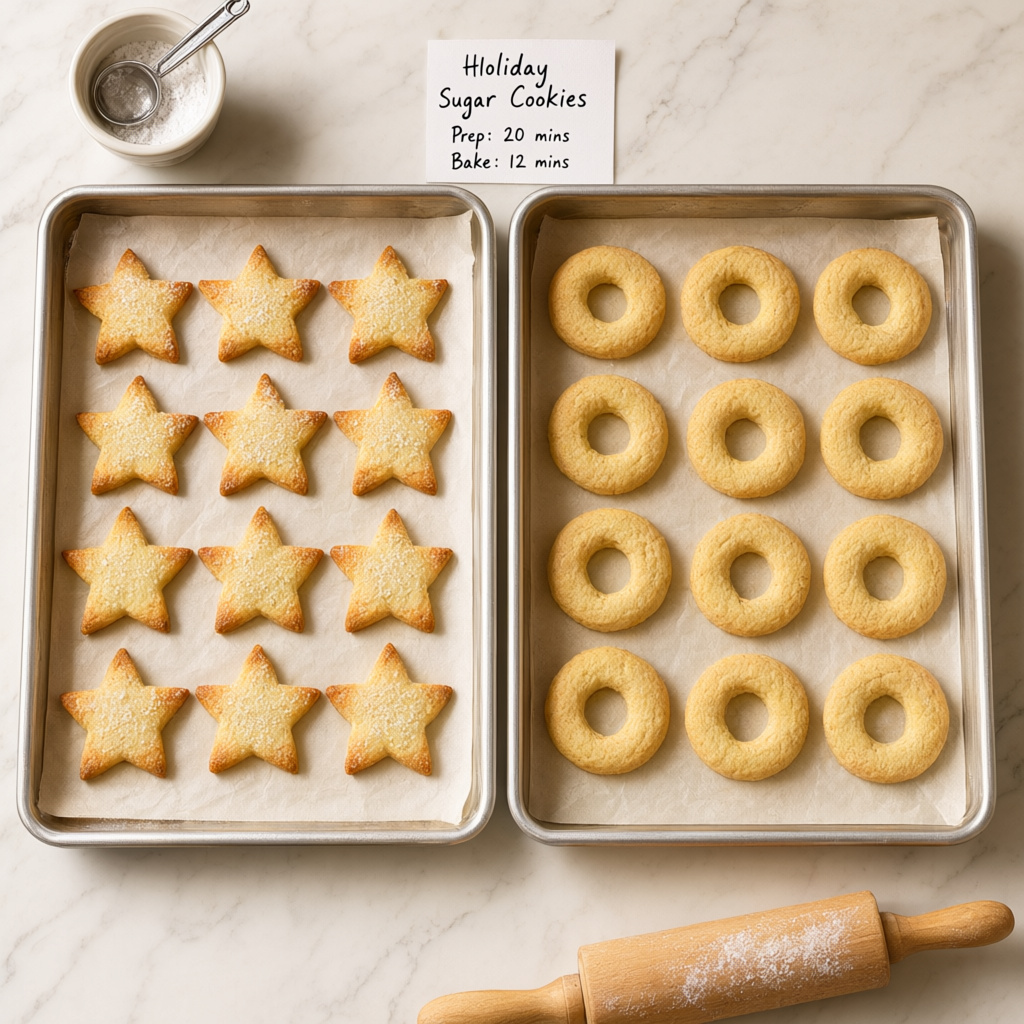} & \caseimg{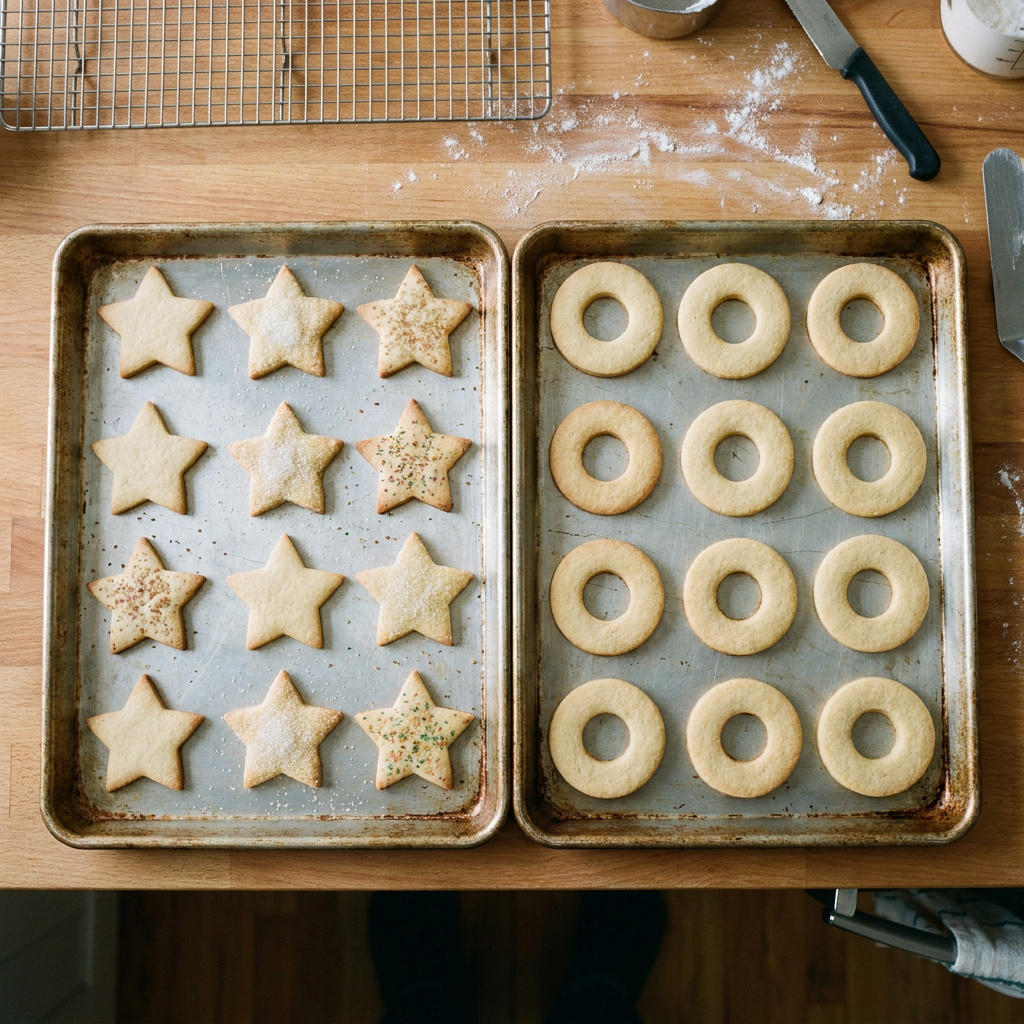} & \caseimg{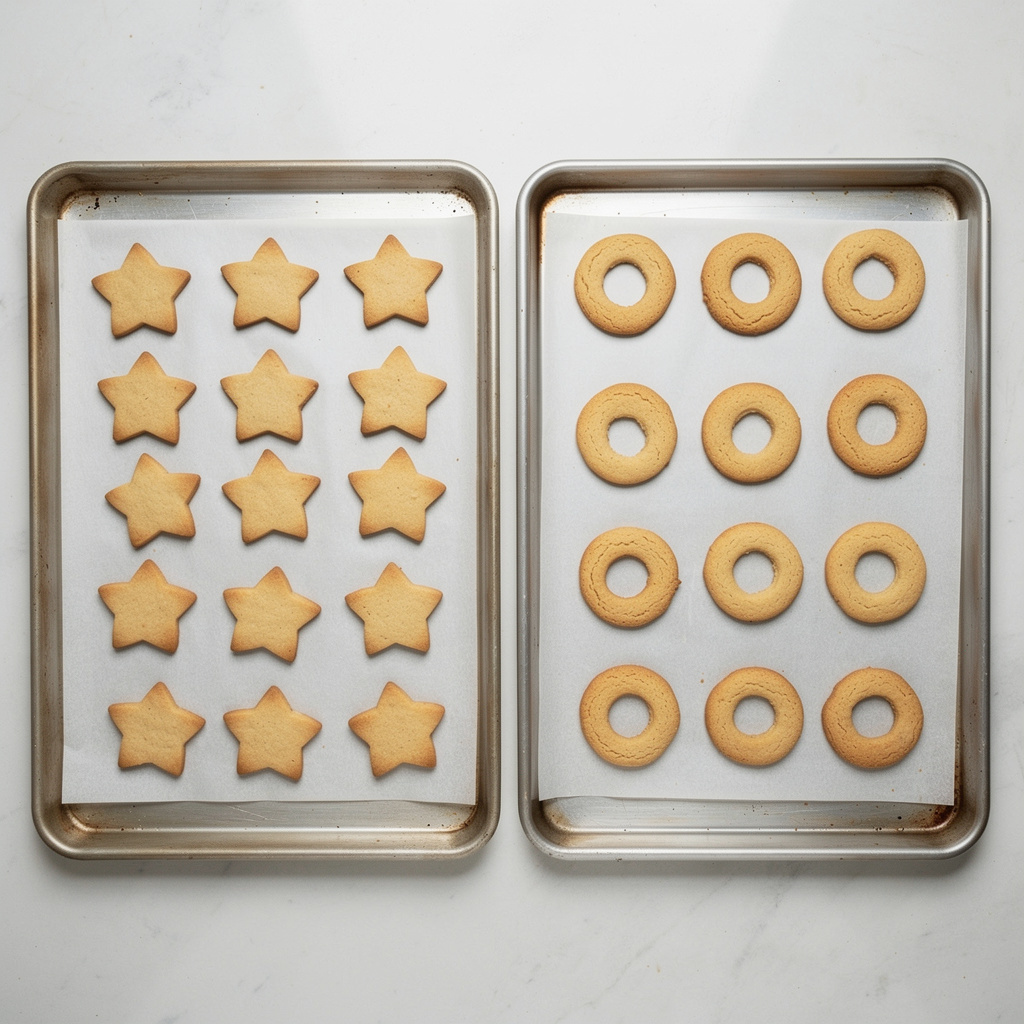} \\
 & \casenote{green!45!black}{Pass}{12 stars; 10 rings} & \casenote{green!45!black}{Pass}{12 stars; 10 rings} & \casenote{red!70!black}{Fail}{12 rings, not 10} & \casenote{red!70!black}{Fail}{12 rings, not 10} & \casenote{red!70!black}{Fail}{15 stars; 12 rings} \\[4pt]
\multicolumn{6}{@{}l@{}}{\scriptsize (b) Blister pack: 3-by-7 cavities; sealed columns by row: none / 1,2,3,4 / all seven.} \\[2pt]
\caseimg{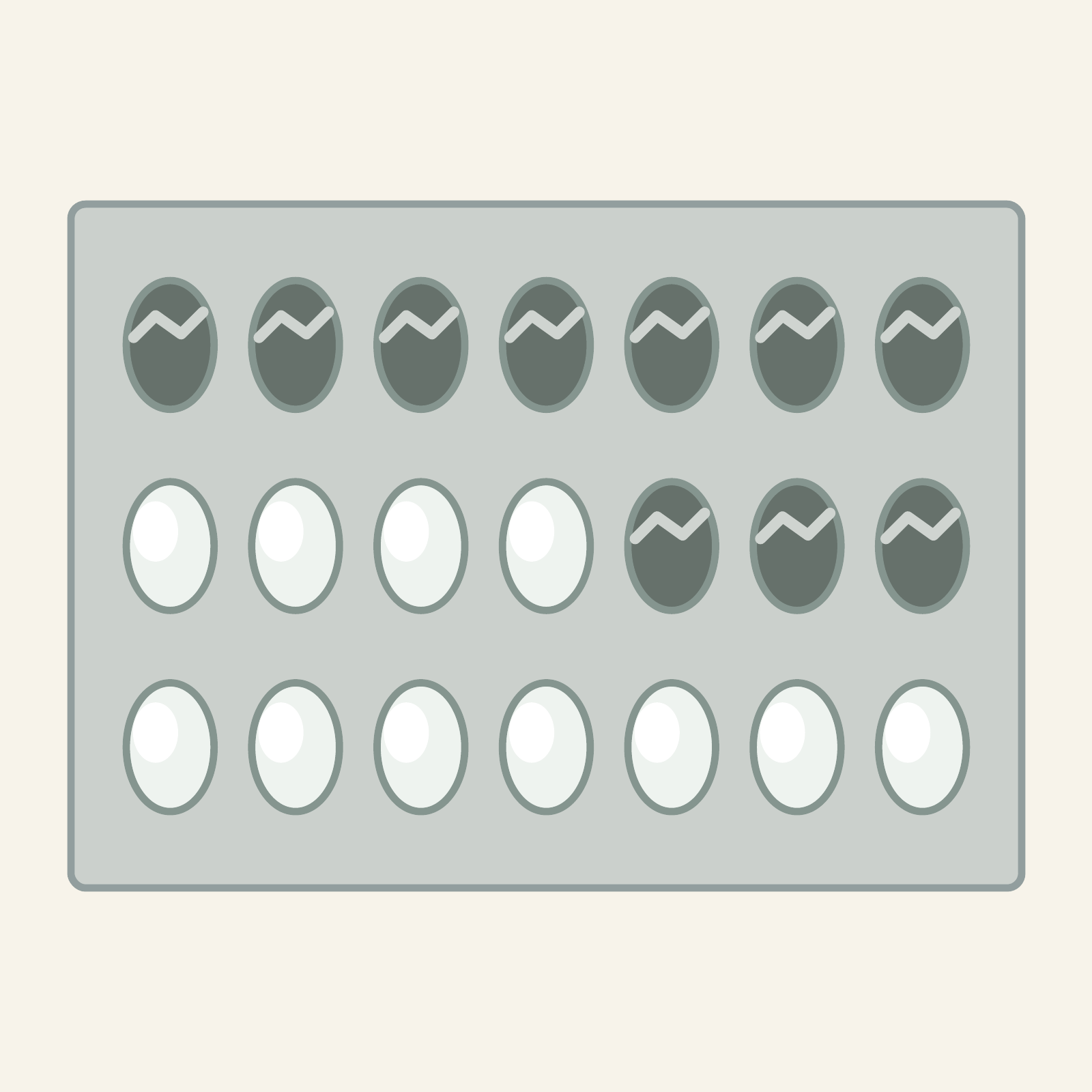} & \caseimg{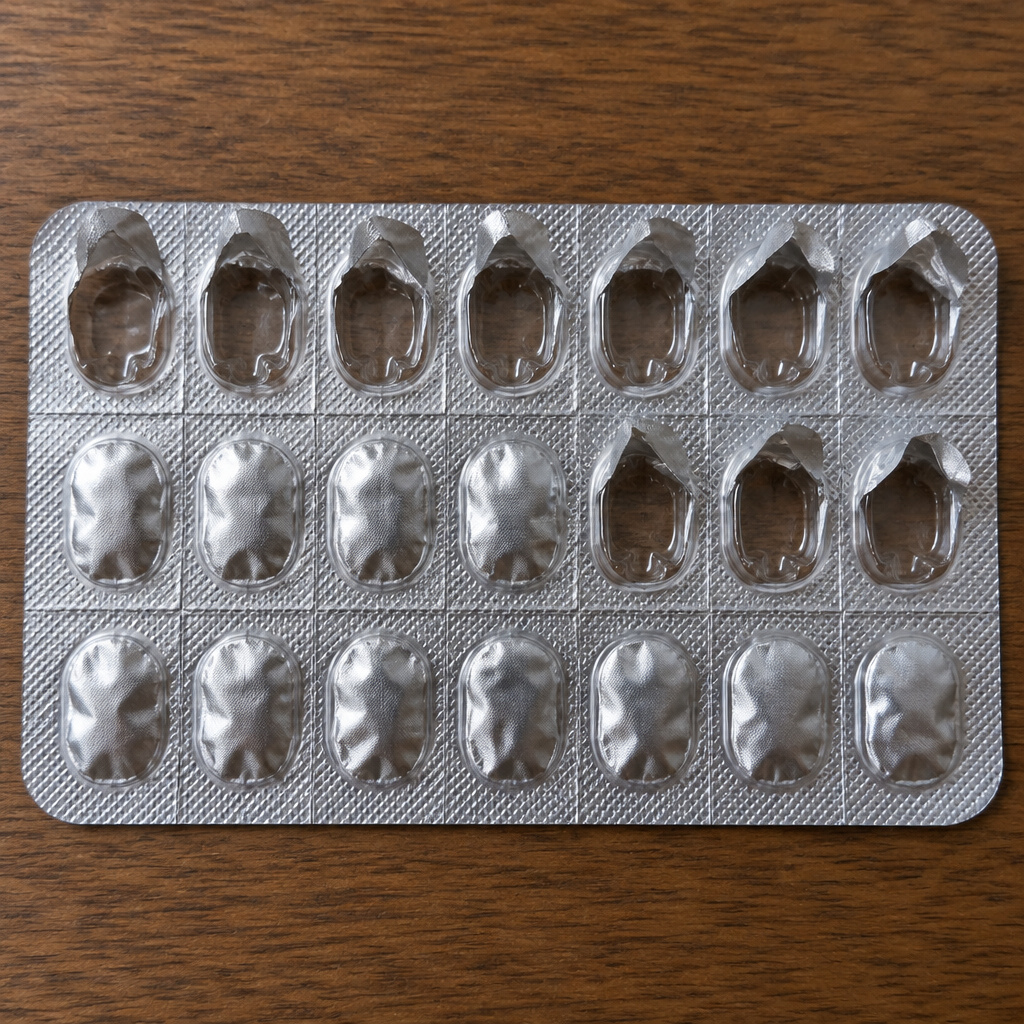} & \caseimg{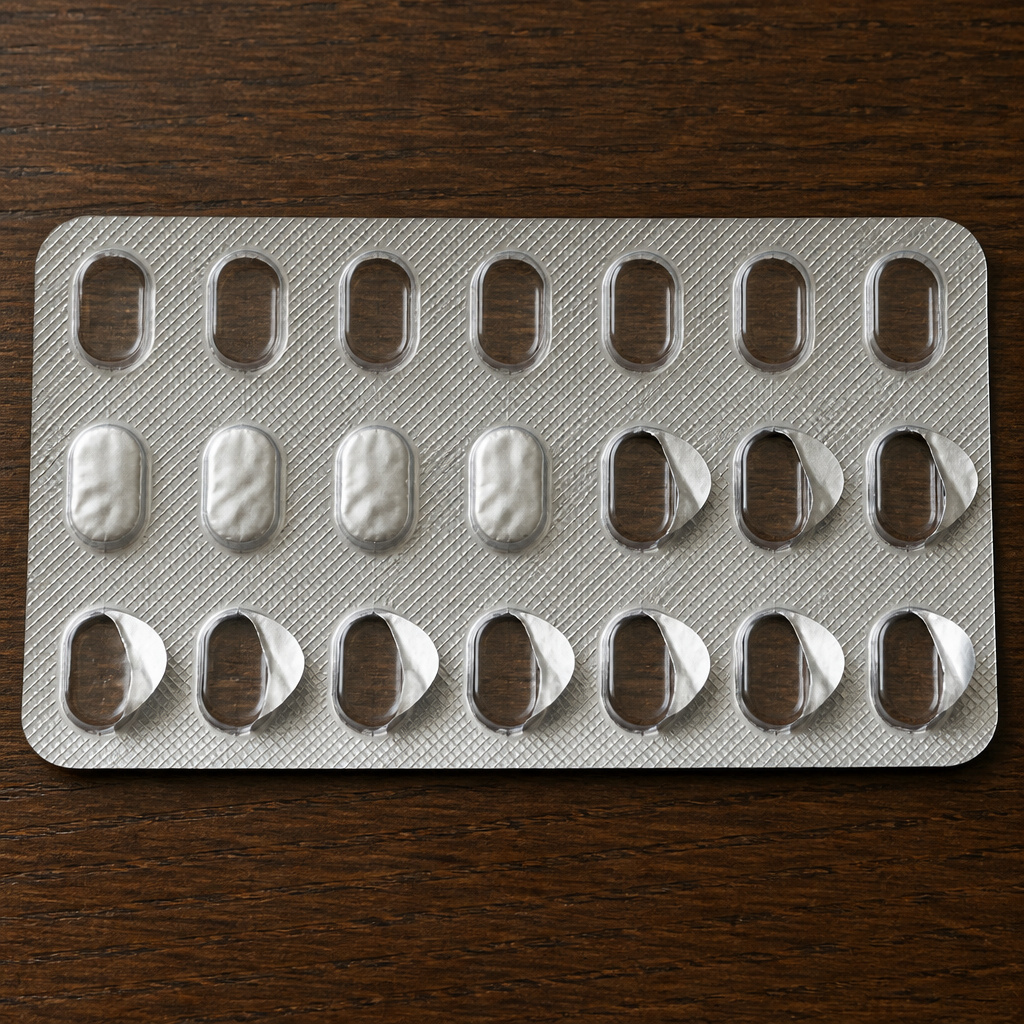} & \caseimg{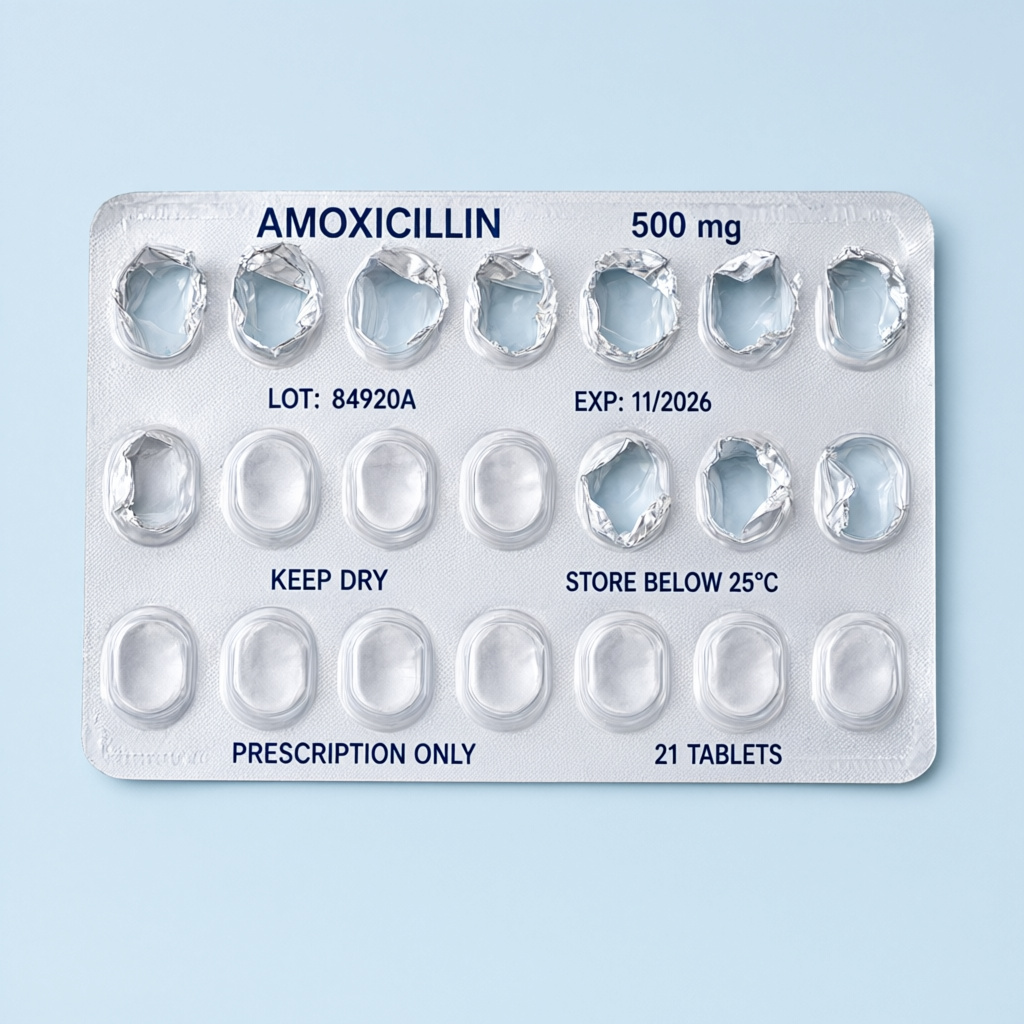} & \caseimg{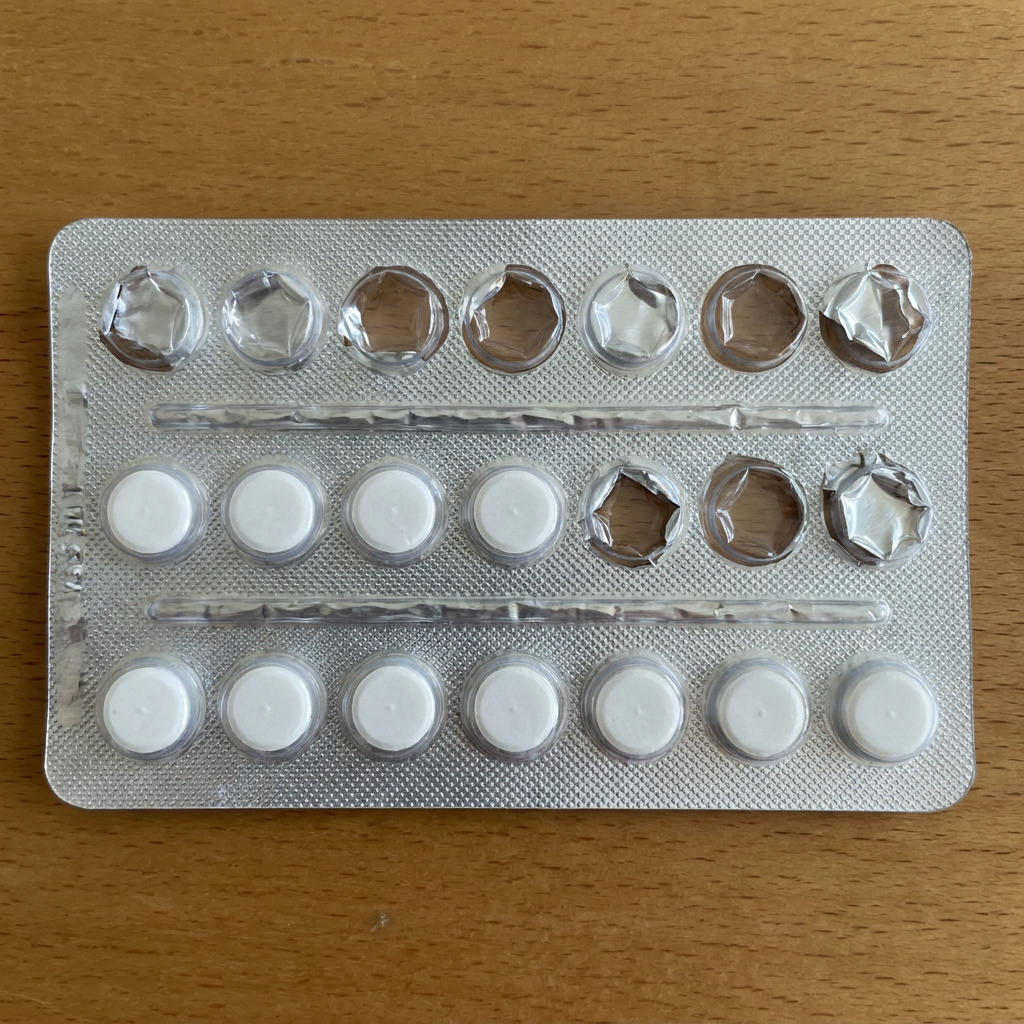} & \caseimg{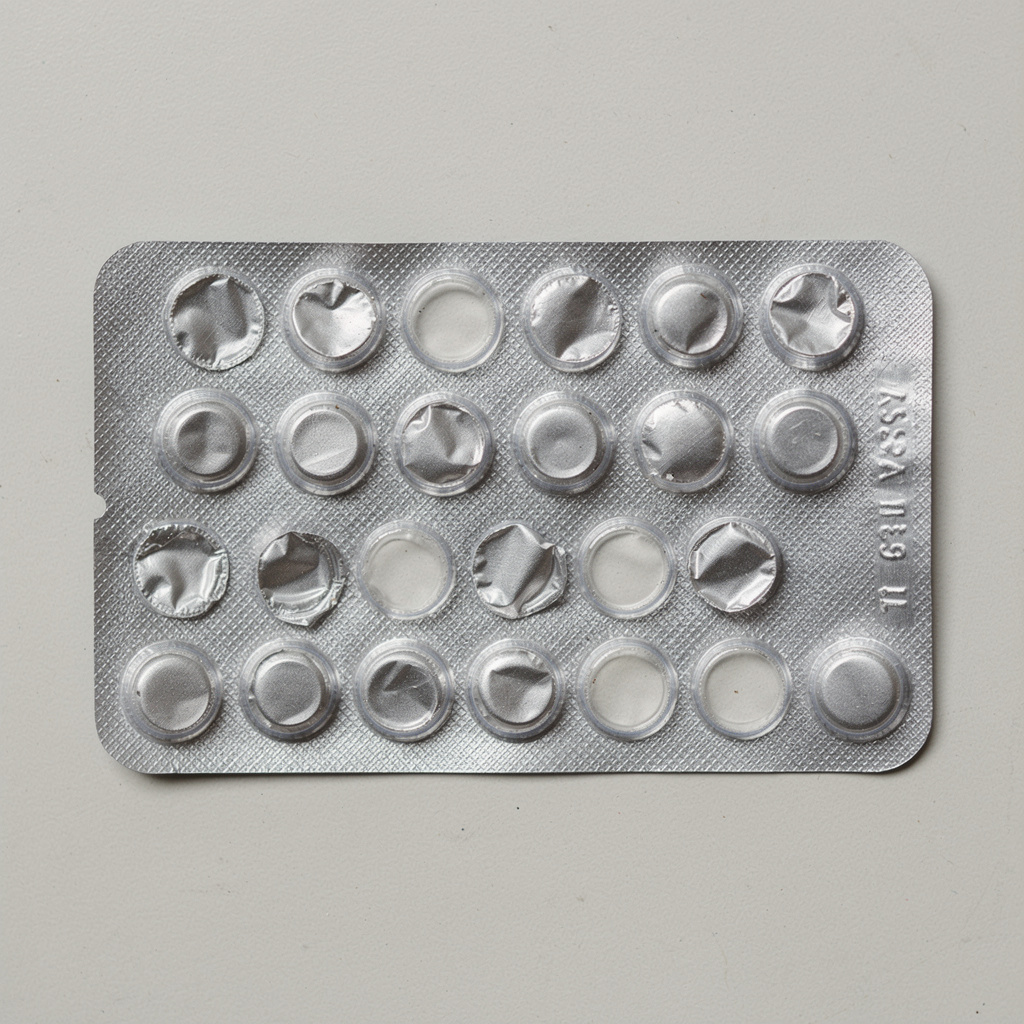} \\
 & \casenote{green!45!black}{Pass}{Specified seals} & \casenote{red!70!black}{Fail}{Bottom row opened} & \casenote{red!70!black}{Fail}{Middle-left seal absent} & \casenote{red!70!black}{Fail}{Top-row seals retained} & \casenote{red!70!black}{Fail}{Four rows, not three} \\[4pt]
\multicolumn{6}{@{}l@{}}{\scriptsize (c) Six key positions: lower-middle cap absent; one shared ellipse, no paint on the exposed base.} \\[2pt]
\caseimg{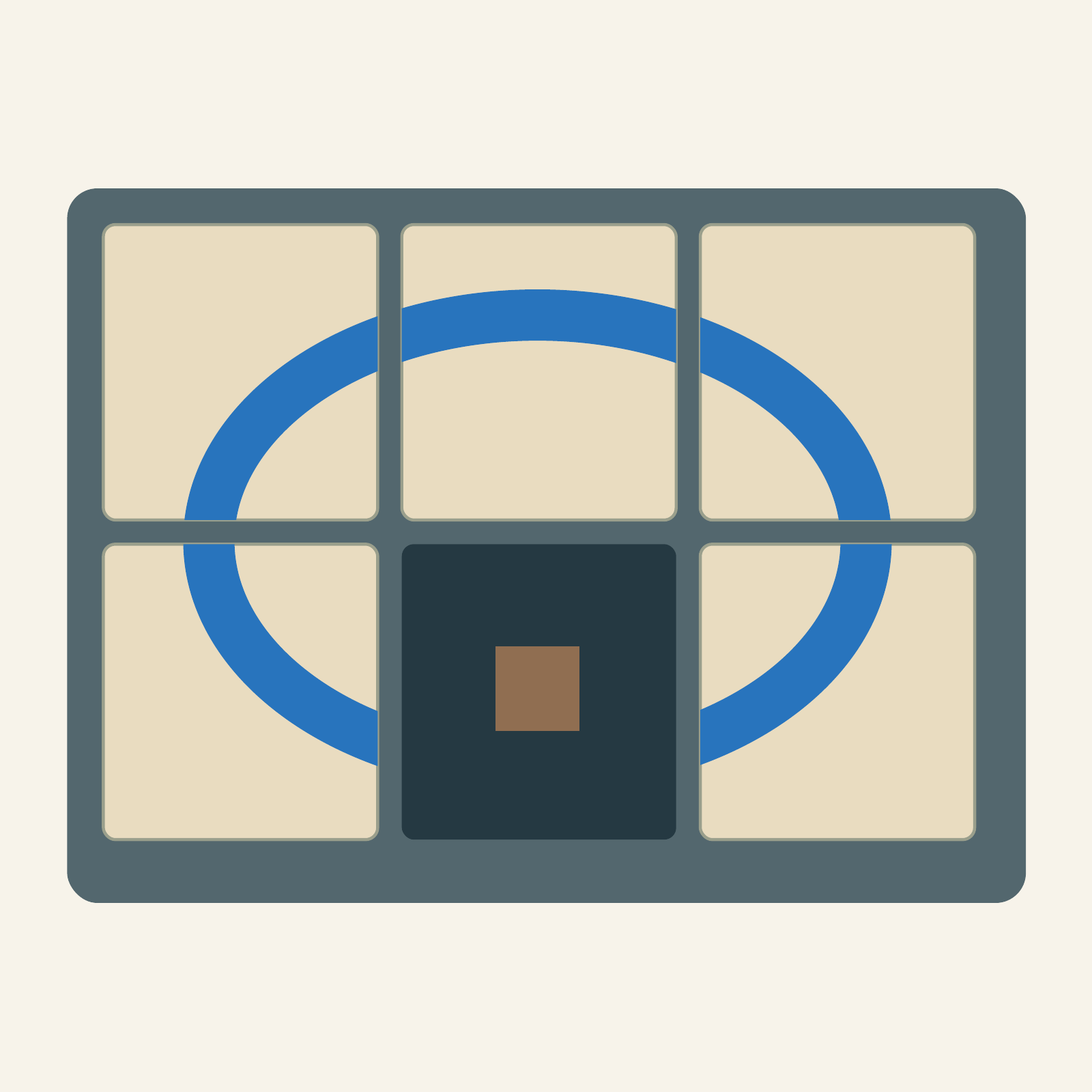} & \caseimg{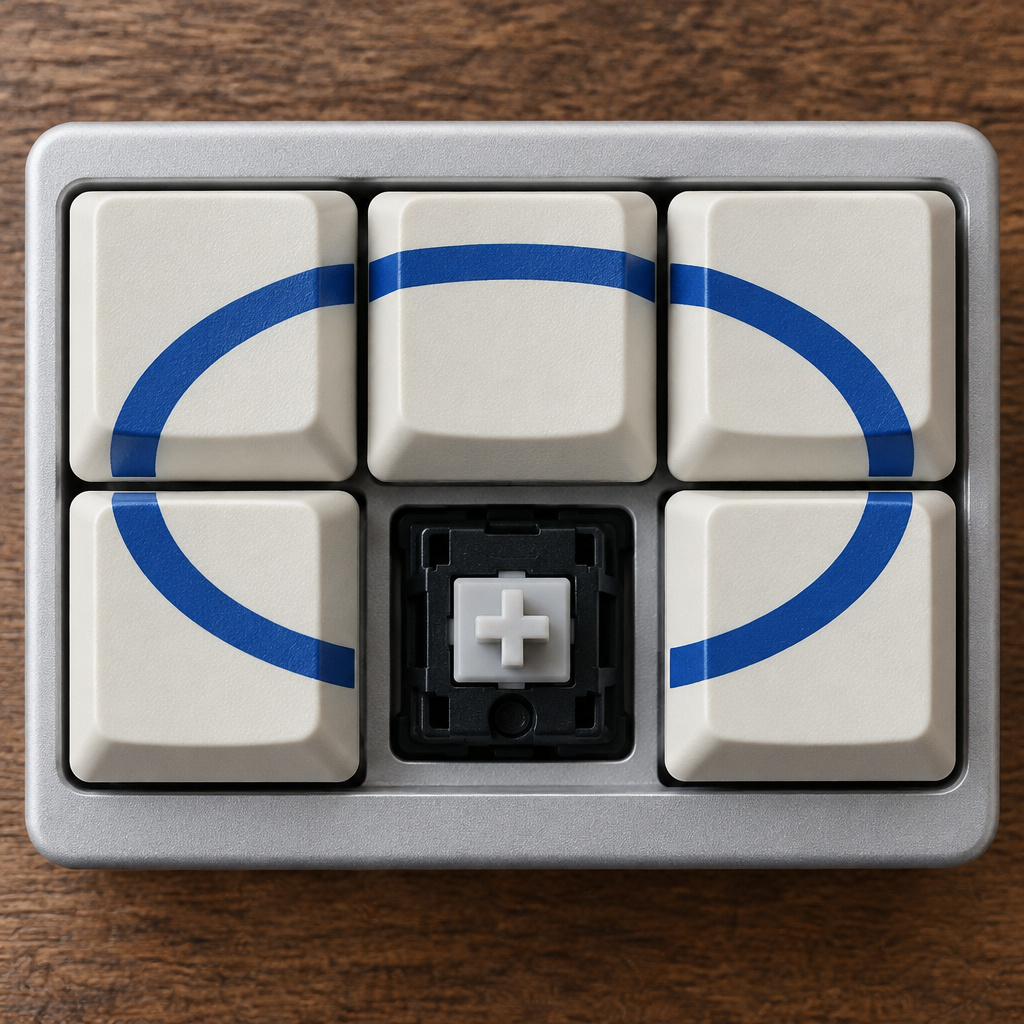} & \caseimg{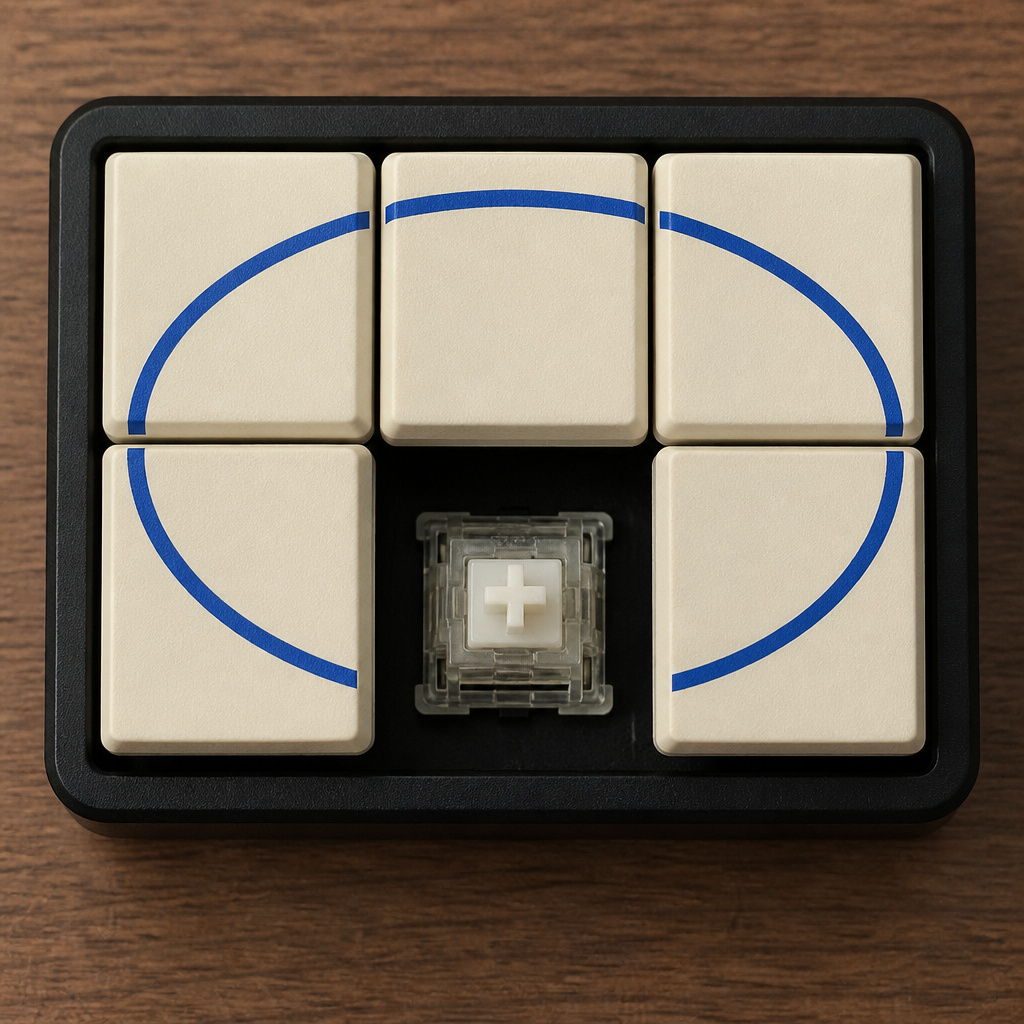} & \caseimg{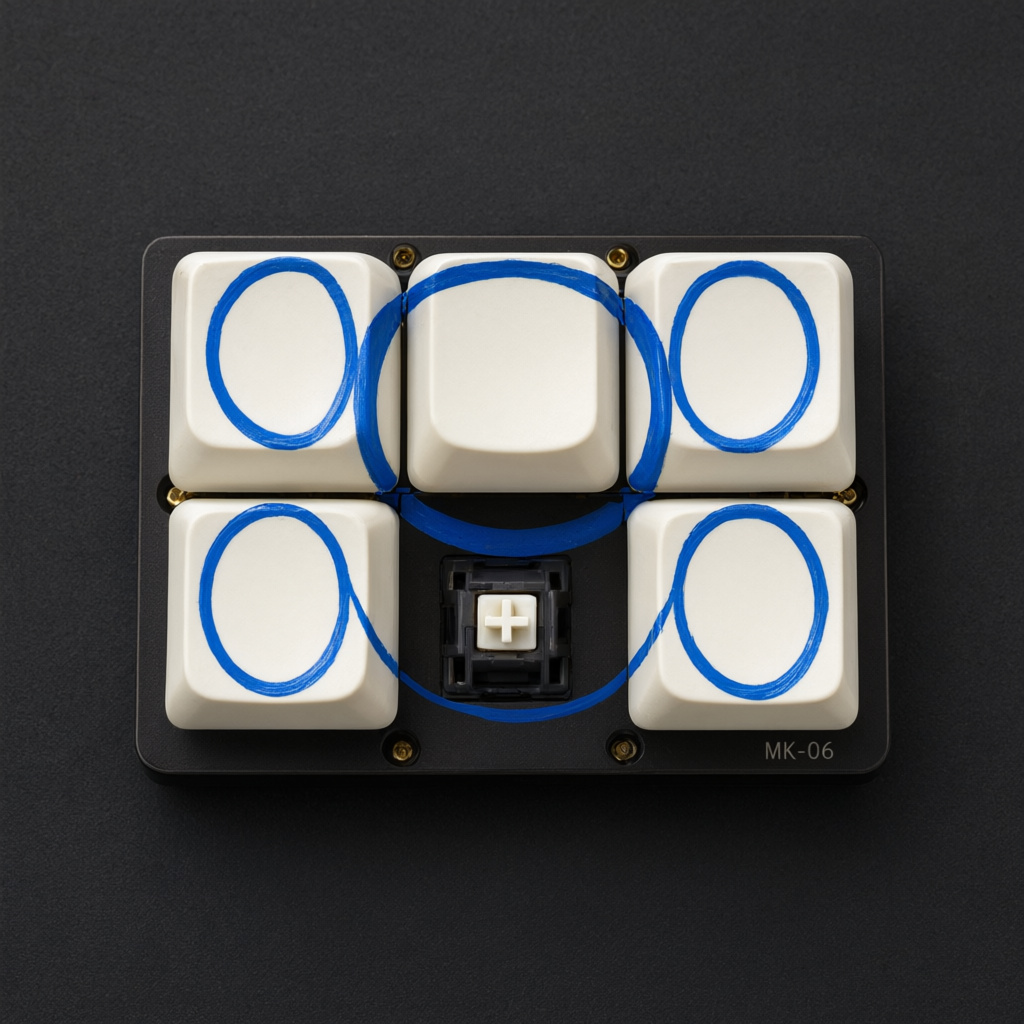} & \caseimg{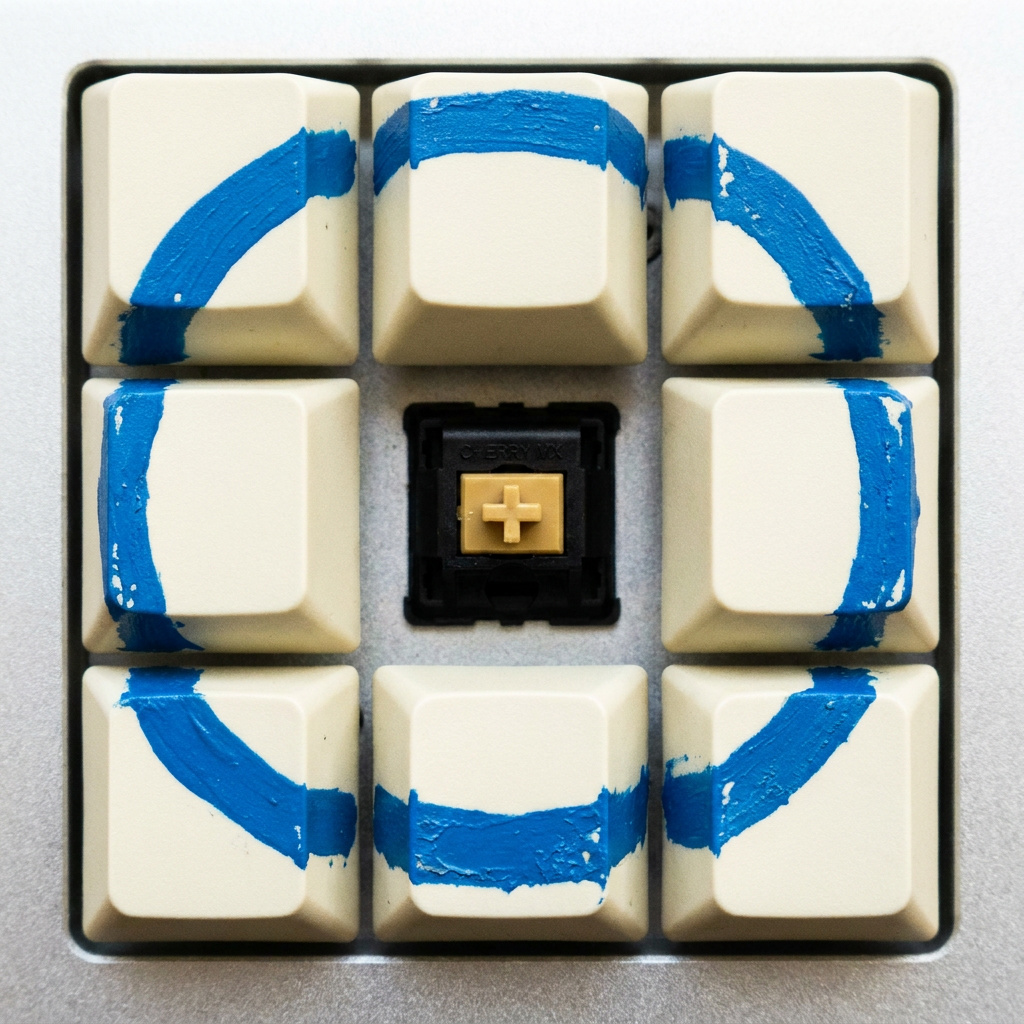} & \caseimg{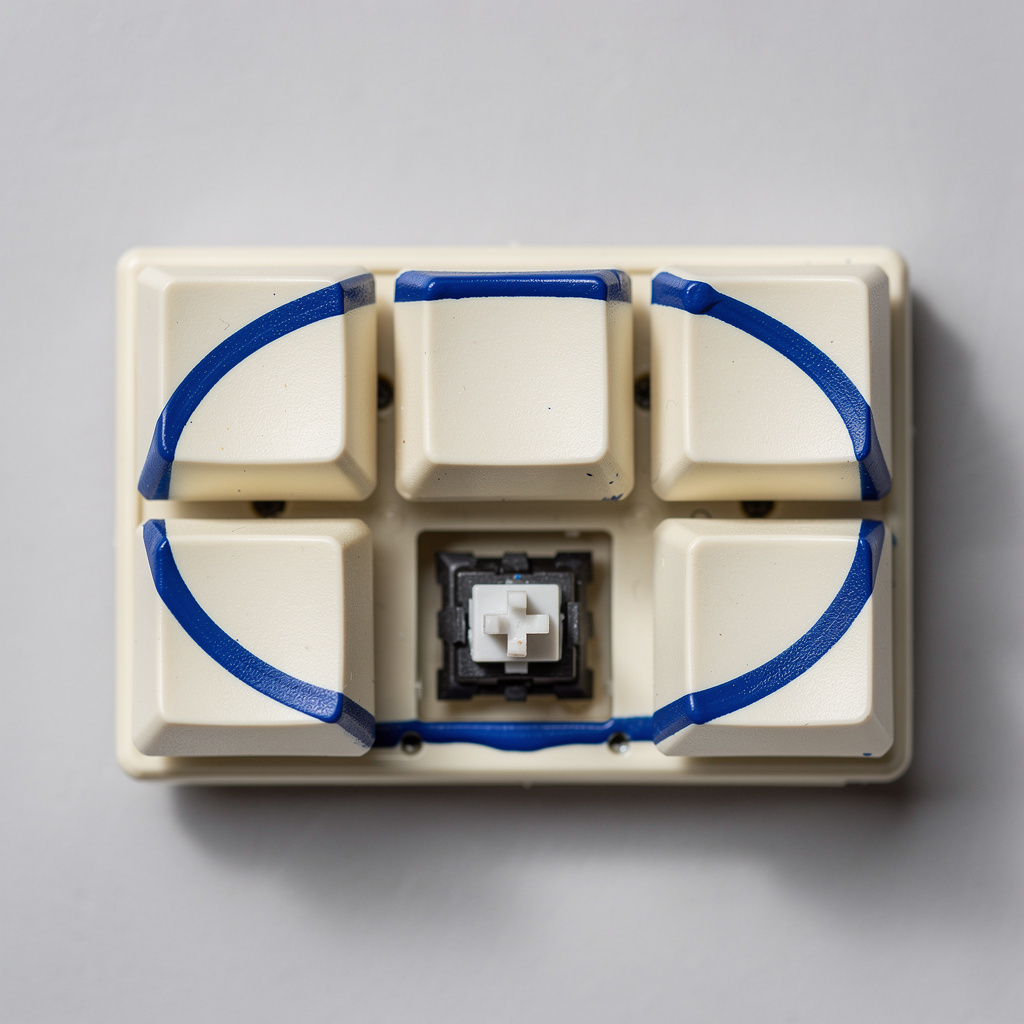} \\
 & \casenote{green!45!black}{Pass}{Shared ellipse clipped} & \casenote{green!45!black}{Pass}{Shared ellipse clipped} & \casenote{red!70!black}{Fail}{Repeated small ellipses} & \casenote{red!70!black}{Fail}{Nine key positions} & \casenote{red!70!black}{Fail}{Paint crosses bare base} \\[4pt]
\multicolumn{6}{@{}l@{}}{\scriptsize (d) Eight panels, clockwise from the top seam: blue, red, yellow, blue, yellow, red, red, yellow.} \\[2pt]
\caseimg{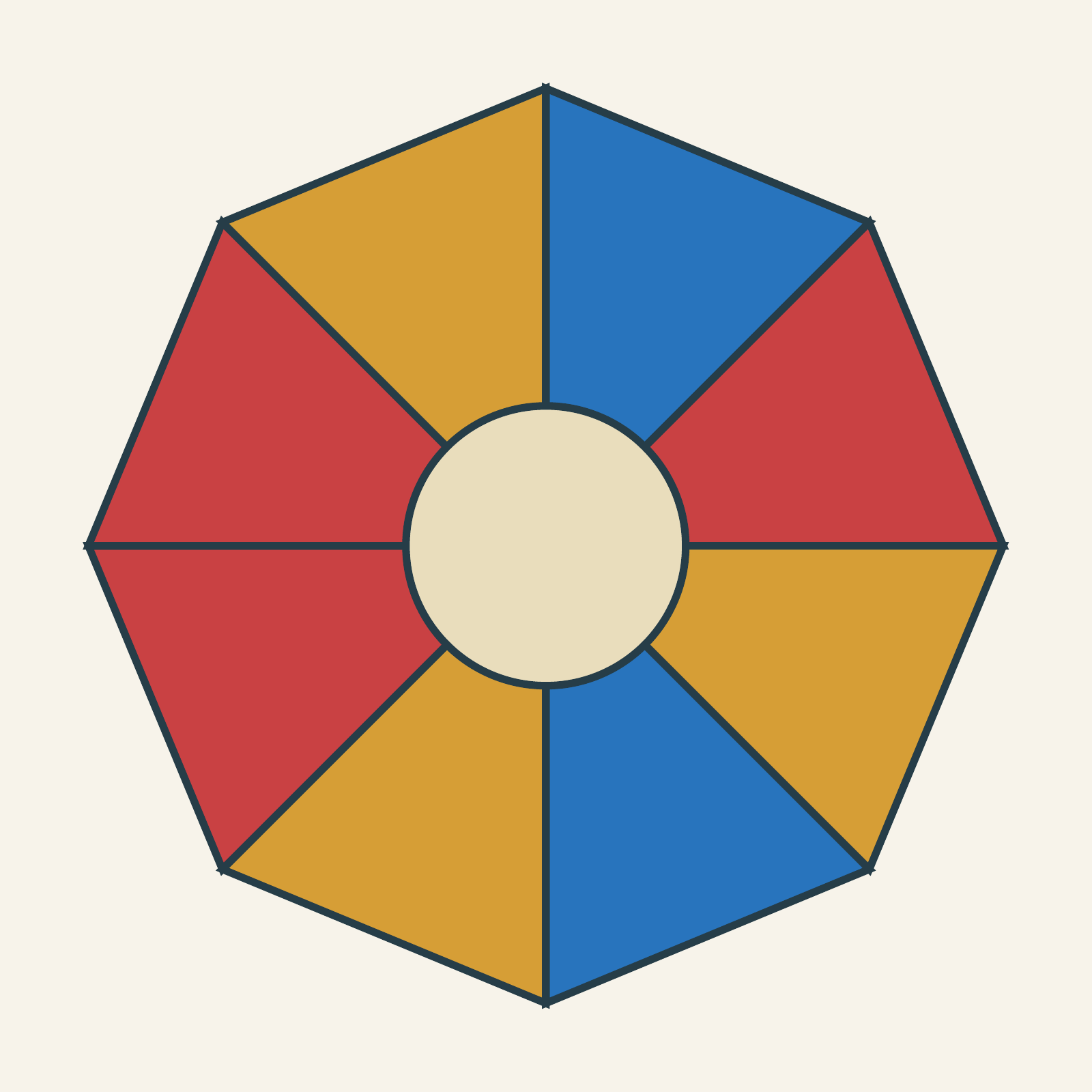} & \caseimg{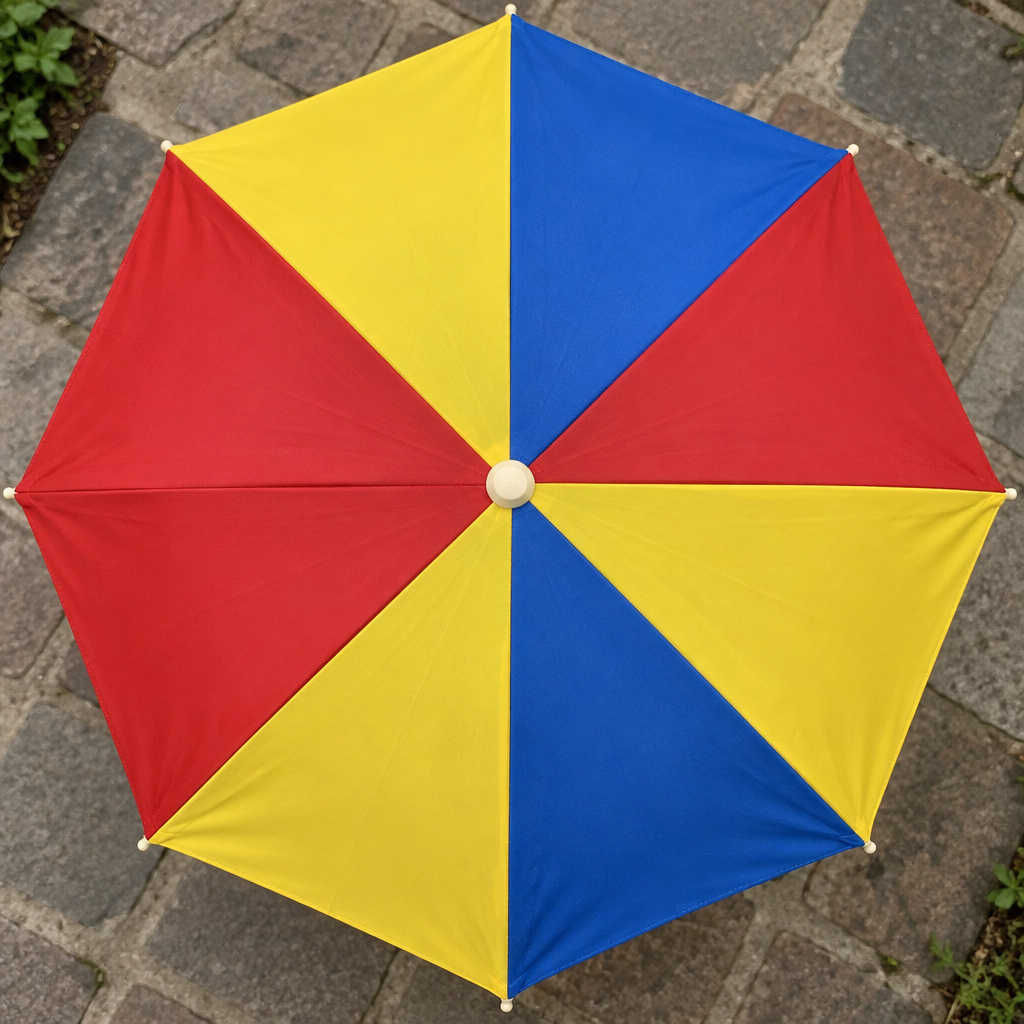} & \caseimg{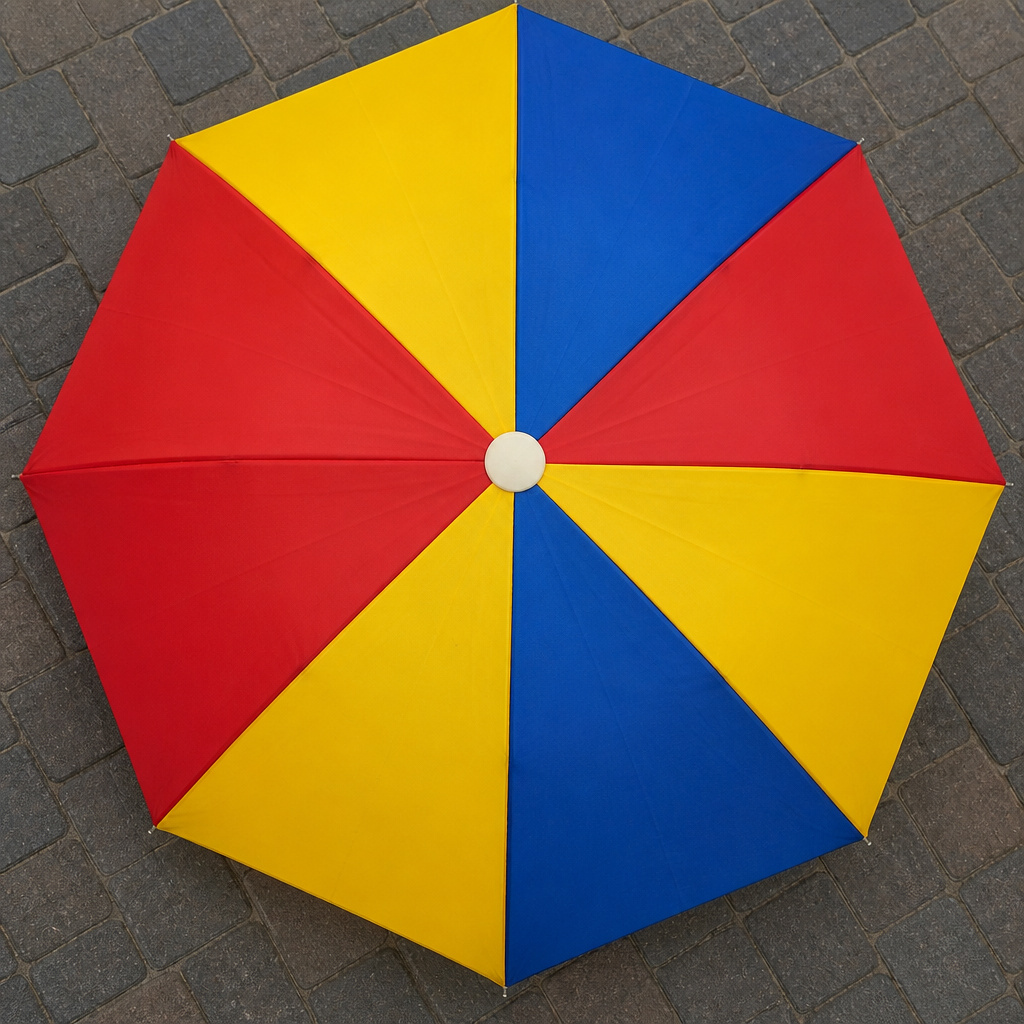} & \caseimg{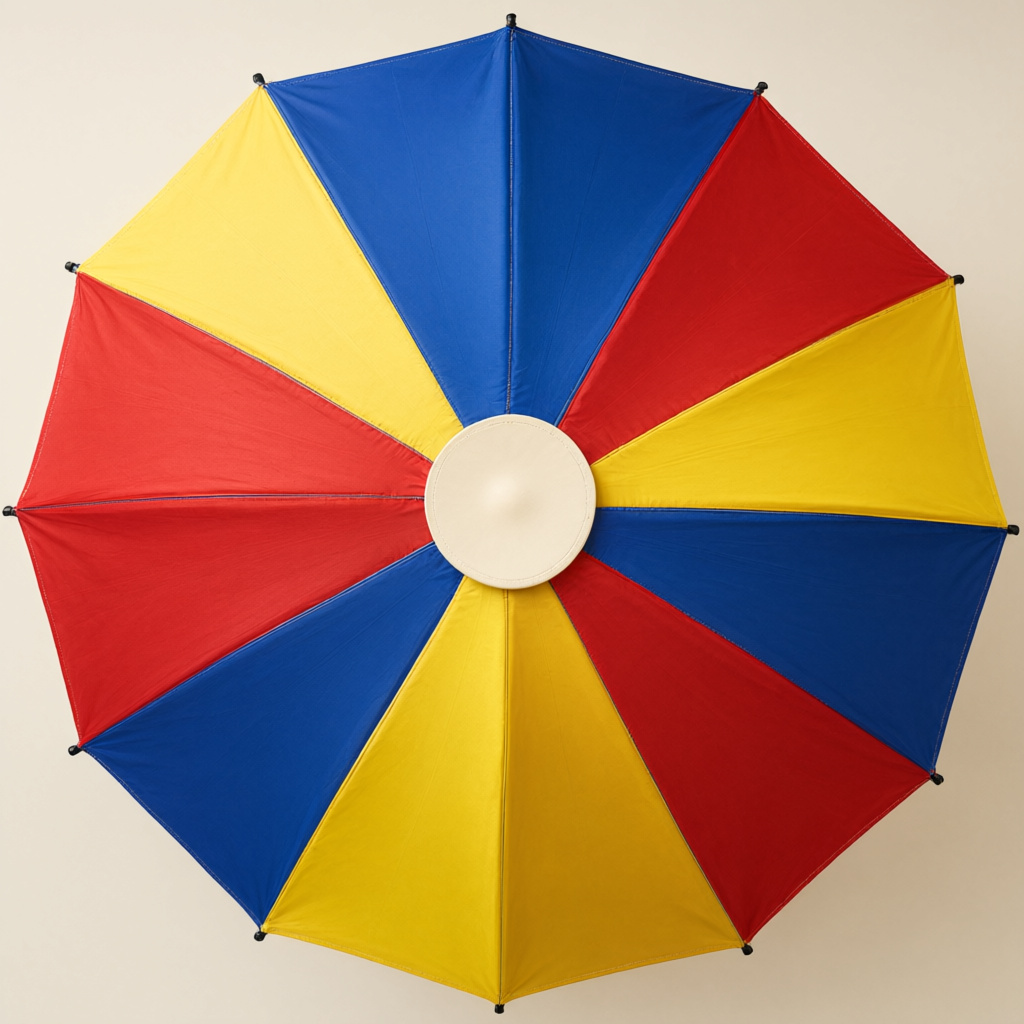} & \caseimg{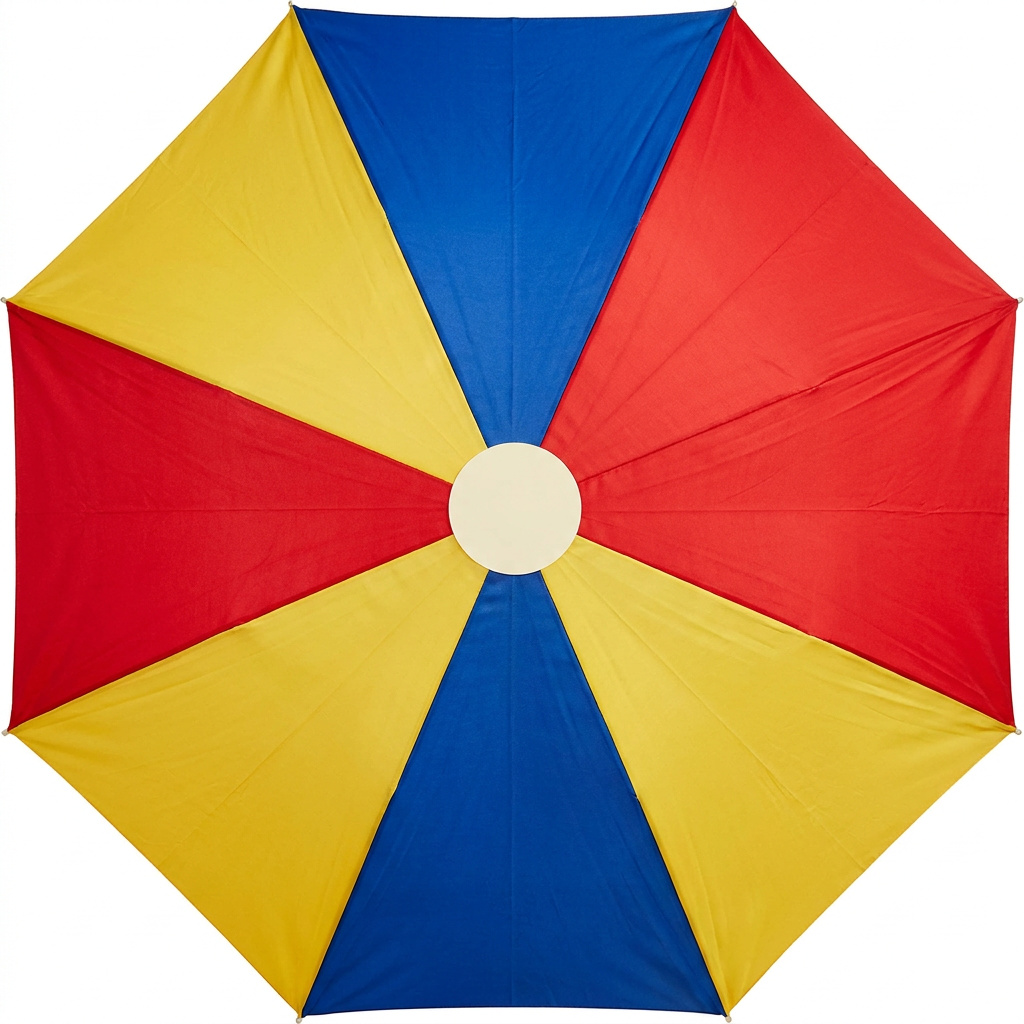} & \caseimg{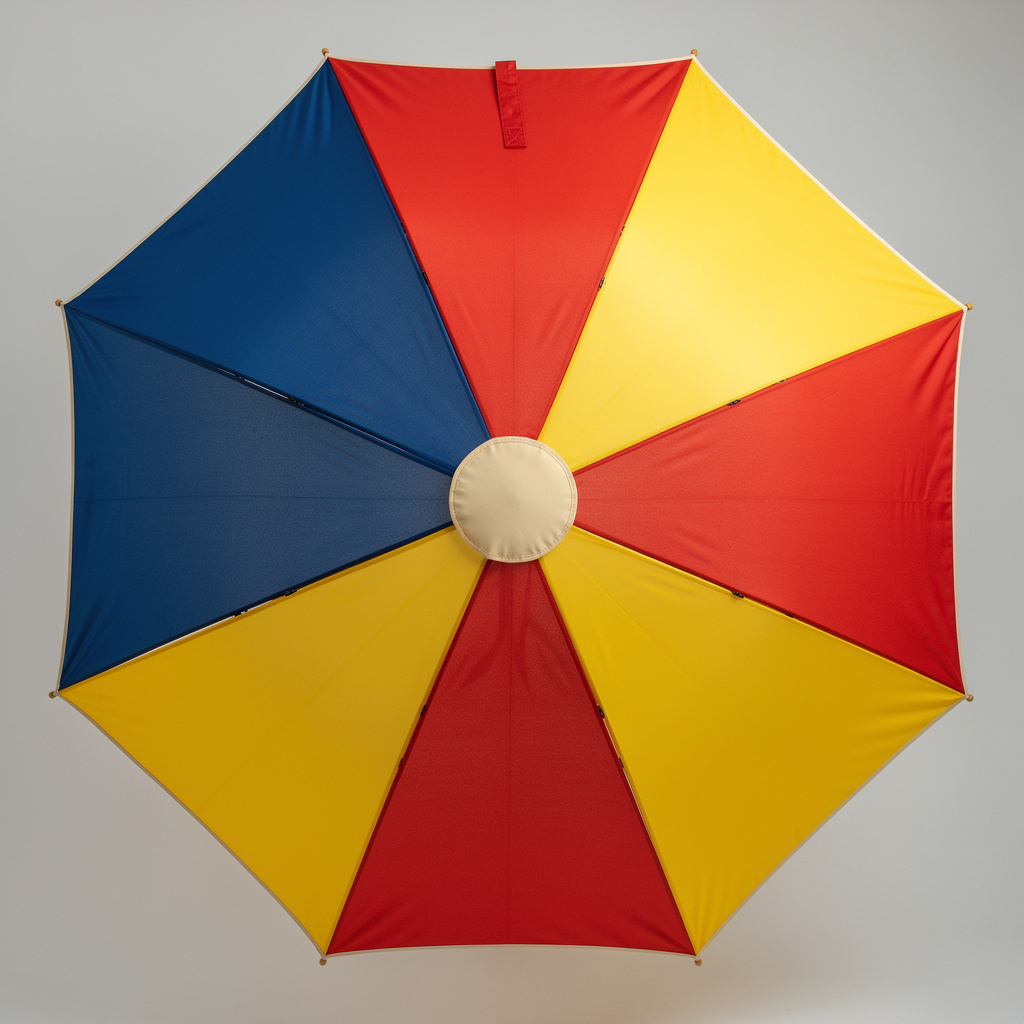} \\
 & \casenote{green!45!black}{Pass}{Specified color order} & \casenote{green!45!black}{Pass}{Specified color order} & \casenote{red!70!black}{Fail}{Ten panels, not eight} & \casenote{red!70!black}{Fail}{Wrong color order} & \casenote{red!70!black}{Fail}{Wrong color order} \\[4pt]
\end{tabular}
\endgroup

\end{figure}

Figure~\ref{fig:qualitative-results} compares five models on four tasks,
with a vector illustration of the target at the left of each row. The
references make the identity requirements visible but are not supplied to
the models. The cases illustrate correct realizations, local errors, and
larger structural departures under the same prompt.

These illustrative cases have one or two passing outputs per row. In the
two-tray task, Sunburst and GPT Image 2 preserve the requested counts and
arrangements; the other three models produce twelve rings instead of ten.
The blister task exposes local state errors: GPT Image 2 opens the bottom
row, Qwen omits the middle-left seal, and Nano Banana Pro retains seals in
the top row. FLUX.2 Max changes the row count.

On the keyboard, Qwen repeats small ellipses, Nano Banana Pro introduces
nine key positions, and FLUX.2 Max extends paint across the exposed base.
For the umbrella, Qwen produces ten panels; Nano Banana Pro and FLUX.2 Max
change the specified color order. The reference encodes the required
identity, not an appearance template; these selected cases do not estimate
error frequencies.

\FloatBarrier
\section{Discussion and Conclusion}

LEVI separates identity requirements from permissible appearance variation.
VIS defines the identities and their equivalence; VVG makes their visual
requirements explicit. Across 1,375 curated tasks and ten model
configurations, the highest whole-task pass rate is 58.69\%. B-VIS
balancing changes the Geometric leader, while component-type counts do
not explain task difficulty. The examples expose concrete errors in
counts, spatial assignments, and local component states despite recognizable
objects. Identity-level evaluation makes these distinctions inspectable.

\paragraph{Limitations.}
Single outputs do not measure generation variability. Scores depend on
post-hoc curation and automatic judgments with task-specific instruction
revisions (Section~\ref{sec:eval-setup}). Domain and construction-batch
imbalance remains after B-VIS reweighting. Whole-task rates cannot isolate
component errors or establish intrinsic domain difficulty; those claims
require matched comparisons and independent human validation.

\FloatBarrier
\clearpage

\bibliography{iclr2027_conference}
\bibliographystyle{iclr2027_conference}

\clearpage
\appendix
\section{Identity Semantics and Task Construction}

\subsection{Identity Spaces and Admissible Answers}
\label{app:vis}
\label{app:projection}

Under the identity rules $s$ defined in Section~\ref{sec:vis}, each point of $\mathcal Z_s$ represents all image realizations of the same identity. A coordinate representation must distinguish these classes and satisfy the constraints defining valid identities. For an ordered symbol sequence, the coordinates record symbol values and order; for a composite identity, they also record any relations needed to bind the components. Equivalent descriptions count as one identity, and legal coordinates may depend on one another.

The permitted answer set $A_s(c)$ follows the prompt. An exact request for $(\triangle,\circ,\square)$ permits one sequence; a request for one of each symbol in any order permits all six permutations. Identity fidelity (Equation~\ref{eq:identity-fidelity}) credits the probability mass of all these permitted identities, including their allowed appearance variations. A vector reference illustrates one valid realization without excluding the others. Images outside the admissible identity domain map to $\bot$; uncertainty in an automatic reader is a separate assessment issue.

\subsection{Composition and Transformations}
\label{app:vis-operations}

Consider finite, nonempty sequences over $\Sigma=\{\triangle,\circ,\square\}$ and the example $z=(\triangle,\circ,\square)$. Table~\ref{tab:identity-operations} distinguishes operations on this example. Composition must preserve the associations between objects: pairing $z$ with a count of the same occurrences requires $n=|z|$. Any spatial or attachment relation that determines the joint identity must also be specified.

\begin{table}[!ht]
\caption{Four operations illustrated with the same symbol sequence.}
\label{tab:identity-operations}
\centering
\small
\renewcommand{\arraystretch}{1.25}
\begin{tabularx}{\textwidth}{@{}p{0.17\textwidth}XX@{}}
\toprule
Operation & Example & Semantic effect \\
\midrule
Composition & Pair $z$ with its count: $(z,3)$ is compatible; $(z,4)$ is not. & Retains component identities and their shared-object constraints. \\
Projection & $P(z)=|z|=3$. & Keeps count; discards symbol values and order. \\
Transformation & Reverse $z$ to $(\square,\circ,\triangle)$. & Changes identity according to a specified rule. \\
Refinement & Replace a count-only specification with an ordered-symbol specification. & Separates identities previously grouped together. \\
\bottomrule
\end{tabularx}
\end{table}
\FloatBarrier

Painting or engraving the same sequence changes its realization when material is not identity-determining. For a structural transformation $\widetilde T$ implementing identity operation $T$, the VVG interpretation $\psi$ must satisfy
\begin{equation}
\psi_{s'}(\widetilde T(g))=T(\psi_s(g)),
\end{equation}
where $g$ is a legal description under $s$, and the transformed description is legal under the target rules $s'$.

\subsection{A Worked Task and the Entropy Decomposition}
\label{app:construction}

\paragraph{From identity to a text-to-image task.}
The following example illustrates how an identity specification determines a text-to-image prompt and its evaluation criteria. It is separate from the evaluated benchmark tasks.

\begin{enumerate}
\item \textbf{Identity.} Three symbol occurrences in the left-to-right order $(\triangle,\circ,\square)$. Color and material may vary while the symbols remain recognizable and distinct.
\item \textbf{VVG description.} Create three typed symbol nodes, assign their values, and connect them in one successor chain. Bind this chain to their left-to-right arrangement on a single plaque. The count is derived from the nodes rather than supplied as an independent value.
\item \textbf{Complete prompt.} ``A front-facing photograph of a single rectangular plaque on a wall. The plaque displays exactly three separate symbols in one horizontal row: a triangle, a circle, and a square, from left to right. All three symbols are fully visible. There are no other symbols or text on the plaque.''
\item \textbf{Evidence and decision.} Identify the plaque and read every symbol on it in left-to-right order, retaining extra or malformed marks for inspection. Accept when the plaque, visibility, and arrangement requirements hold and the observed sequence is exactly $(\triangle,\circ,\square)$. A changed material is permitted; an extra symbol or swapped pair violates the request.
\end{enumerate}

The generator receives the prompt, not the VVG description or vector reference. A legal description may express a different sequence, so grammar membership alone does not establish task success. Likewise, a rendering preserves the intended identity only if its pixels retain the specified facts and object associations. The automatic evaluation procedure is described in Section~\ref{sec:eval-setup}.

\paragraph{Entropy and correctness.}
\label{app:distribution}
For a fixed prompt $C=c$, let $X\sim q_\theta(\cdot\mid c)$ and $Z_s=\phi_s(X)$. Since $\phi_s$ is deterministic, $H(Z_s\mid X,C=c)=0$. The chain rule therefore gives
\begin{equation}
\begin{aligned}
H(X,Z_s\mid C=c)&=H(X\mid C=c)\\
&=H(Z_s\mid C=c)+H(X\mid Z_s,C=c).
\end{aligned}
\label{eq:entropy-proof}
\end{equation}
The two terms distinguish identity uncertainty from appearance variation within an identity. Entropies use base-two logarithms and discrete representations; continuous quantities require a declared quantization. For identity coordinates, dependence is retained through
\begin{equation}
H(Z_1,\ldots,Z_k\mid C=c)
=\sum_{j=1}^{k}H(Z_j\mid Z_{<j},C=c),
\label{eq:coordinate-chain}
\end{equation}
where $Z_{<j}$ denotes the preceding coordinates. A count derived from a complete object list adds zero conditional uncertainty. Always generating one wrong identity gives zero identity entropy and zero fidelity; generating only permitted identities can give positive entropy and perfect fidelity. The empirical whole-task pass rate measures success under the chosen task set and assessment procedure.

\clearpage
\raggedbottom
\makeatletter
\newenvironment{leviSupplementFigure}{%
  \par\addvspace{10pt}\noindent\begin{minipage}{\linewidth}%
  \def\@captype{figure}%
}{\end{minipage}\par\addvspace{10pt}}
\newenvironment{leviSupplementTable}{%
  \par\addvspace{8pt}\noindent\begin{minipage}{\linewidth}%
  \def\@captype{table}%
}{\end{minipage}\par\addvspace{8pt}}
\makeatother
\section{Task Types and Representative Examples}
\label{app:task-examples}

The tasks in Figure~\ref{fig:supp-gallery-overview} require specific relationships between visual elements: outlined versus solid letters, relative heights, endpoint connections, object positions, regional counts, and additions between stages. They use different physical media to express these requirements. Recognizing the object category or matching a total count is insufficient when these relationships are incorrect.

\begin{leviSupplementFigure}
\centering
\includegraphics[width=\textwidth]{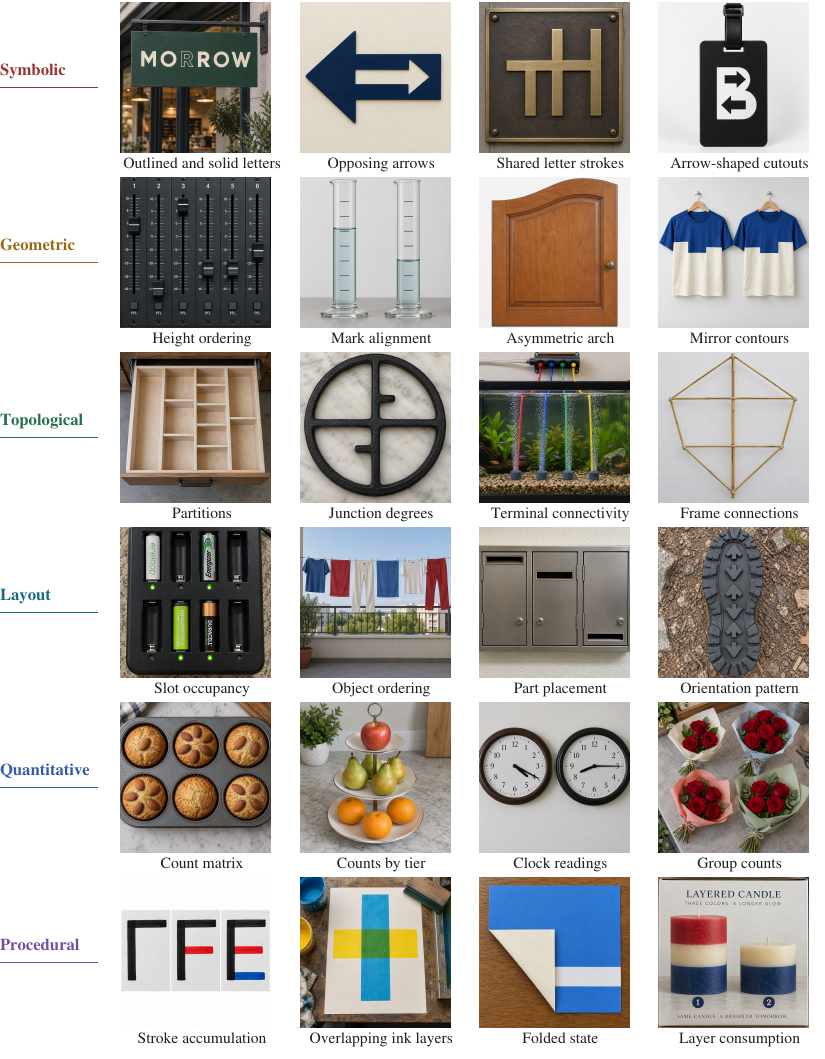}
\caption{Twenty-four tasks across six identity domains. All images are uncropped GPT Image 2.5 Sunburst outputs. Labels describe the identity requirement. The examples illustrate task diversity; they are not a sample for estimating error frequency.}
\label{fig:supp-gallery-overview}
\end{leviSupplementFigure}
\newpage

\section{Prompt and Model Output Analysis}

\label{app:prompt-analysis}

We compare five models on six tasks distinct from Appendix~\ref{app:task-examples} and Figure~\ref{fig:qualitative-results}. Each case presents the prompt, required visual evidence, an identity schematic, and model outputs. Schematics were not generator inputs. PASS/FAIL labels are automatic judgments. The cases emphasize visible failures rather than estimate error frequencies.

\subsection{Symbolic: note positions and notehead types}

\paragraph{Prompt.}

\begin{quote}\small A frontal close photograph of a large-print practice score on a piano music stand. A single five-line staff carries six separated notes left to right. Counting staff lines upward from the bottom, place their noteheads on the bottom line, in the first space, on the third line, in the third space, on the fifth line, and in the fourth space. The second and fifth heads are hollow; the others are filled. Each has its own upward stem, without flags, beams, text or extra notes. Use fine staff lines and large clear noteheads.\end{quote}

\paragraph{Identity evidence.} Count staff lines from bottom to top. The six noteheads must occupy line 1, space 1, line 3, space 3, line 5, and space 4. Only the second and fifth heads are hollow. Each note has an upward stem without a flag or beam.

\begin{leviSupplementFigure}

\centering

\includegraphics[width=\textwidth]{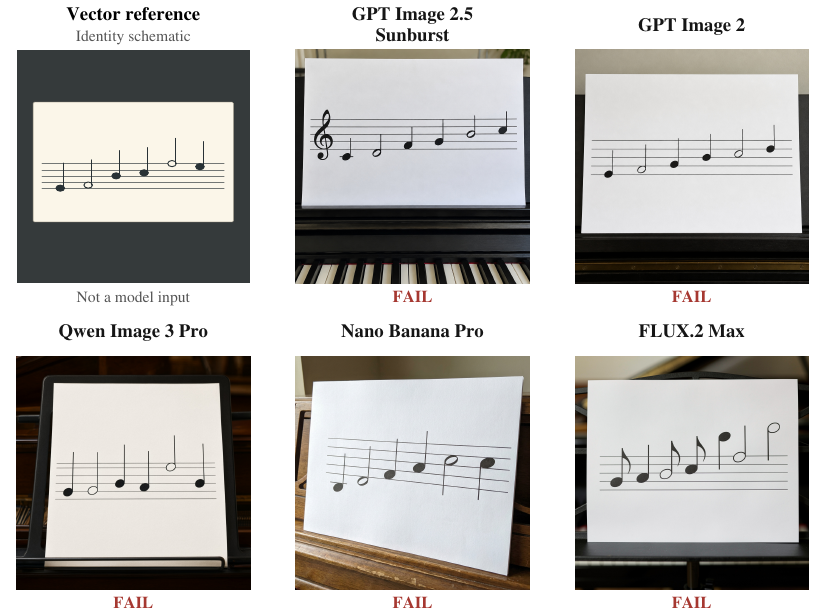}

\caption{Symbolic: note positions and notehead types. The schematic illustrates the required identity; labels show automatic judgments.}

\label{fig:supp-case-music}

\end{leviSupplementFigure}

\paragraph{Model comparison.} All five models receive FAIL. Sunburst places the first note below the staff; GPT Image 2 places the third note on the second line rather than the third. Qwen draws six staff lines. Nano Banana Pro gives the last two notes downward stems. FLUX adds a seventh note and flags. Musical appearance alone does not preserve the specified notation.

\newpage

\subsection{Geometric: a rotated photograph with a level horizon}

\paragraph{Prompt.}

\begin{quote}\small A frontal photograph of an upright rectangular tabletop picture frame. Its outer edges are horizontal and vertical. Inside the wide frame, a rectangular seascape photograph is clearly rotated clockwise, with all four corners remaining inside the frame. Do not tilt the whole frame or turn the picture into a trapezoid. The seascape contains a broad clear sea–sky horizon that remains horizontal in image coordinates, parallel to the outer frame bottom.\end{quote}

\paragraph{Identity evidence.} Compare three orientations separately: the upright outer frame, the clockwise-rotated rectangular photograph, and the horizontal sea--sky boundary inside it. All four photograph corners must lie inside the frame. The horizon must remain parallel to the outer frame bottom even though the photograph is rotated.

\begin{leviSupplementFigure}

\centering

\includegraphics[width=\textwidth]{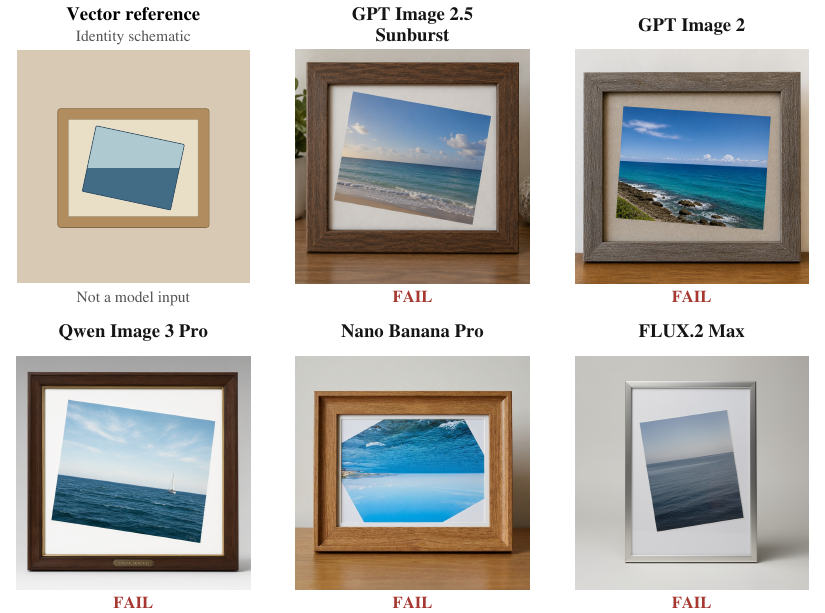}

\caption{Geometric: a rotated photograph with a level horizon. The schematic illustrates the required identity; labels show automatic judgments.}

\label{fig:supp-case-horizon}

\end{leviSupplementFigure}

\paragraph{Model comparison.} All five models receive FAIL. Sunburst, GPT Image 2, and Qwen tilt the horizon downward to the right along with the photograph. Nano Banana Pro replaces the rectangular photograph with a polygonal shape. FLUX rotates the photograph counterclockwise and tilts its horizon upward to the right. The failure concerns the relationship between orientations, not whether the scene resembles a framed seascape.

\newpage

\subsection{Topological: alternating over-under crossings}

\paragraph{Prompt.}

\begin{quote}\small An overhead close photograph of a flat canvas tote bag with large appliqued fabric letters on its front. Two separate dark navy letters, E on the left and F on the right, have distinct red vertical ribbons. At E, the red ribbon passes behind, in front of, behind the horizontal strokes in top-to-bottom order. At F, the red ribbon passes in front of, behind the horizontal strokes in top-to-bottom order. The broad ribbon edges make every crossing legible. The tote has no other visible lettering.\end{quote}

\paragraph{Identity evidence.} Trace each red ribbon through its own letter. At E, the ribbon passes behind the top bar, in front of the middle bar, and behind the bottom bar. At F, it passes in front of the top bar and behind the middle bar. Each crossing must show the specified layer order while the ribbon remains continuous.

\begin{leviSupplementFigure}

\centering

\includegraphics[width=\textwidth]{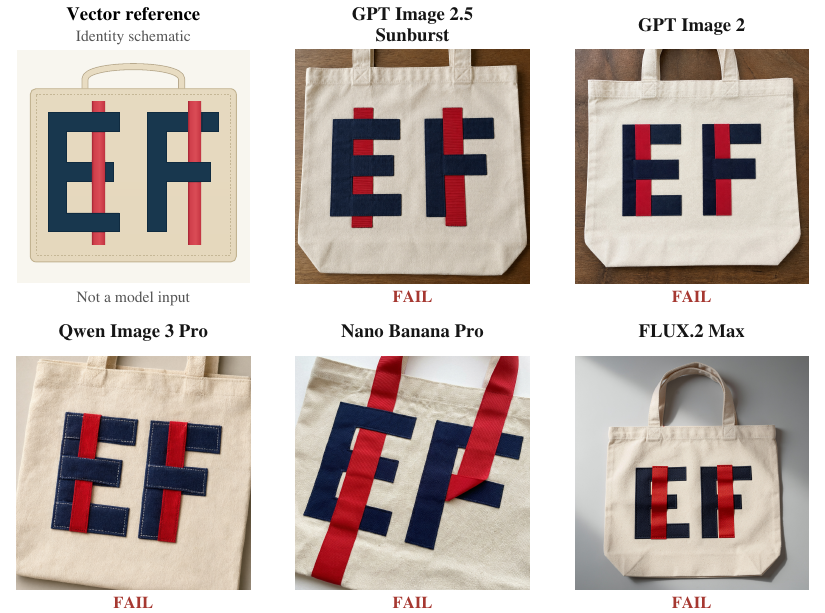}

\caption{Topological: alternating over-under crossings. The schematic illustrates the required identity; labels show automatic judgments.}

\label{fig:supp-case-ribbons}

\end{leviSupplementFigure}

\paragraph{Model comparison.} All five models receive FAIL. Sunburst places the ribbon behind all three bars of E. Qwen reverses the required E sequence, while FLUX keeps red visible over every bar. Nano Banana Pro leaves the ribbon in front of the bottom E bar. GPT Image 2 joins some letter bars to ribbon edges instead of forming crossings. The defining relation is which material covers the other at each intersection.

\newpage

\subsection{Layout: rotating a window pattern}

\paragraph{Prompt.}

\begin{quote}\small A frontal architectural close photograph of two square stained-glass window sashes side by side. Each has three rows of three square panes. On the left, each pane may have no diagonal lead bar, a diagonal joining its upper-left and lower-right corners, or one joining its lower-left and upper-right corners. Use both diagonal directions and at least one pane with no diagonal. The right sash is a clockwise quarter-turn counterpart of the actual left lead-bar layout, including pane locations and diagonal directions. The glass is a single light amber color so that all dark lead lines are easy to distinguish. No pane has two diagonals or additional decorative lines.\end{quote}

\paragraph{Identity evidence.} Each window must contain a three-by-three grid of square panes. Read each pane as empty or containing one of the two diagonals. Rotating the left pattern clockwise by a quarter turn changes both pane positions and diagonal directions. The right window must match that transformed pattern; the schematic shows one valid pair.

\begin{leviSupplementFigure}

\centering

\includegraphics[width=\textwidth]{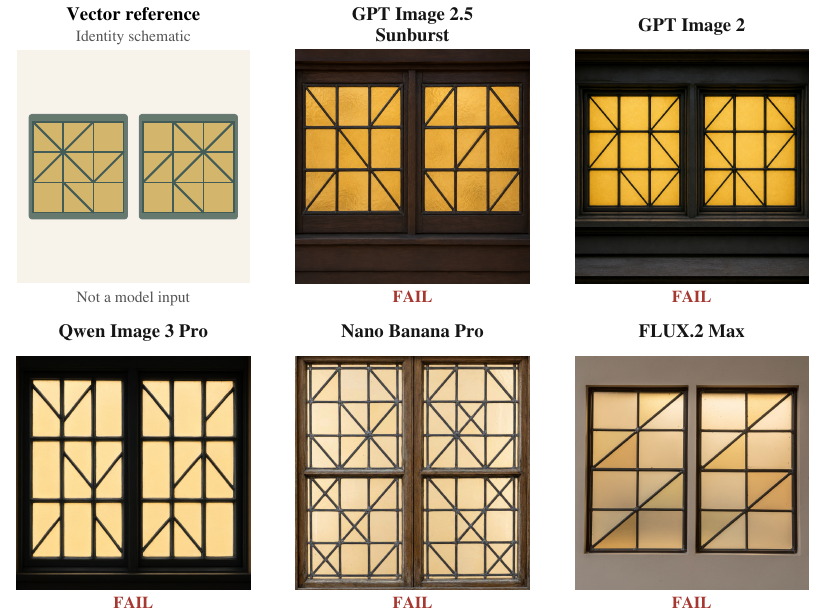}

\caption{Layout: rotating a window pattern. The schematic illustrates the required identity; labels show automatic judgments.}

\label{fig:supp-case-windows}

\end{leviSupplementFigure}

\paragraph{Model comparison.} All five models receive FAIL. Sunburst and GPT Image 2 place diagonals in directions inconsistent with a clockwise quarter turn. Qwen draws bars that end midway along pane edges. Nano Banana Pro adds stacked grids and crossed diagonals. FLUX uses two columns and four rows and repeats the left pattern. Matching the window style does not establish the required rotation.

\newpage

\subsection{Quantitative: counts in four wing regions}

\paragraph{Prompt.}

\begin{quote}\small Overhead photograph of one large butterfly-shaped decorated cookie, head upward, four clear wing lobes and a narrow central body. All wings have pale-blue icing with separate black dots: image upper-left two, upper-right four, lower-left one, lower-right three. Dots stay clear of wing edges and body. Preserve the independent counts rather than enforcing bilateral symmetry. Only this cookie on the plate, no text.\end{quote}

\paragraph{Identity evidence.} Orient the butterfly with its head upward and count dots within each wing separately. The upper-left, upper-right, lower-left, and lower-right targets are 2, 4, 1, and 3. Each dot must remain inside its wing and separate from the body. The requested counts are asymmetric even though the butterfly has a familiar bilateral outline.

\begin{leviSupplementFigure}

\centering

\includegraphics[width=\textwidth]{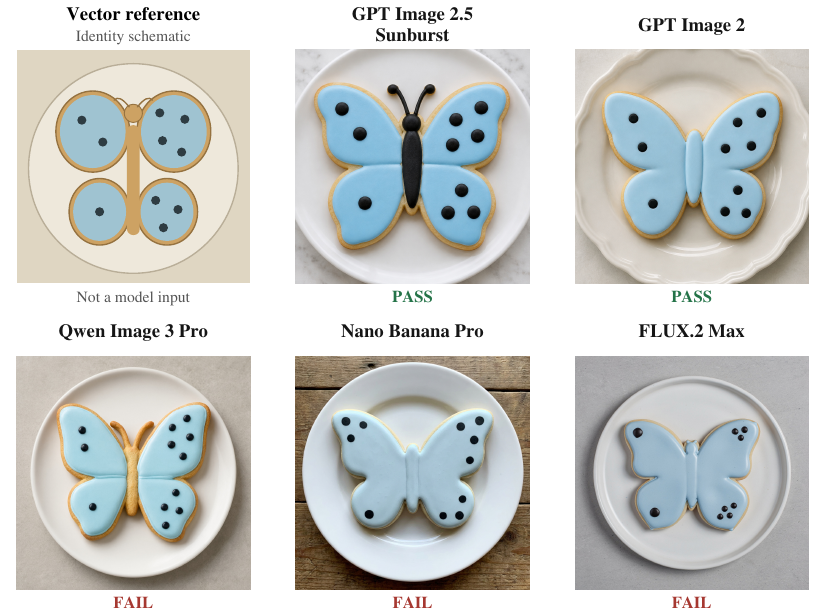}

\caption{Quantitative: counts in four wing regions. The schematic illustrates the required identity; labels show automatic judgments.}

\label{fig:supp-case-butterfly}

\end{leviSupplementFigure}

\paragraph{Model comparison.} Sunburst and GPT Image 2 preserve all four counts and receive PASS. Qwen places five dots on the upper-right wing and four on the lower-right, rather than four and three. Nano Banana Pro adds a third dot to the upper-left wing. FLUX produces counts of 1, 3, 1, and 4. The errors concern specific regions; a plausible butterfly shape does not satisfy their independent counts.

\newpage

\subsection{Procedural: a partially piped heart}

\paragraph{Prompt.}

\begin{quote}\small Overhead photograph of a cake during piping. One broad ivory icing line forms only the left half of a heart: from the central upper notch around the upper-left lobe to the bottom point. The right half is absent. A piping nozzle meets that bottom endpoint, with no extra icing tail. No dotted heart, arrows or completed exemplar.\end{quote}

\paragraph{Identity evidence.} Trace one icing line from the upper central notch, around the left lobe, to the bottom point. The right half must remain absent. The piping nozzle must meet the bottom endpoint, with no icing tail beyond it. The image specifies the current visible stage of the piping process.

\begin{leviSupplementFigure}

\centering

\includegraphics[width=\textwidth]{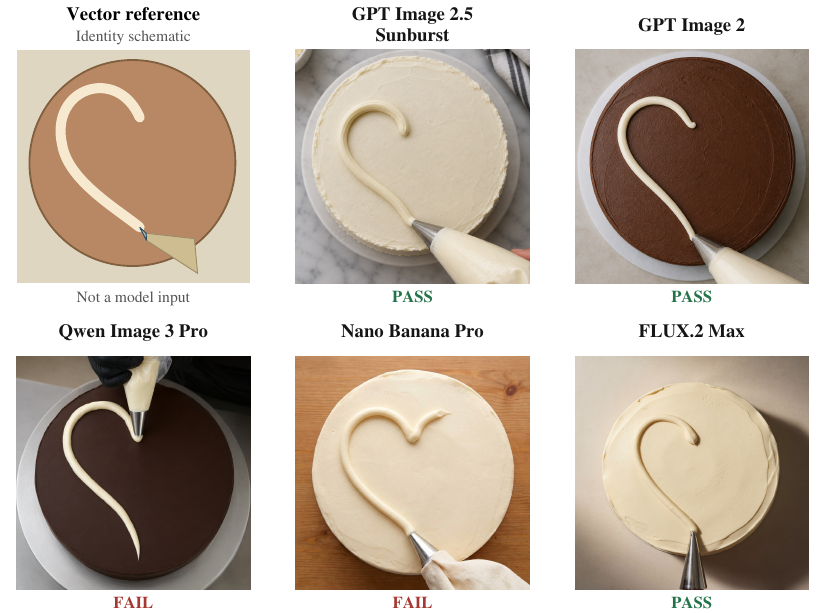}

\caption{Procedural: a partially piped heart. The schematic illustrates the required identity; labels show automatic judgments.}

\label{fig:supp-case-piping}

\end{leviSupplementFigure}

\paragraph{Model comparison.} Sunburst, GPT Image 2, and FLUX show the left half of the heart with the nozzle at the bottom point and receive PASS. Qwen places the nozzle at the upper notch instead. Nano Banana Pro extends the icing to the right of that notch, adding part of the missing half. These failures distinguish the location of the active endpoint from the extent of the completed stroke.

\section{Extended Experimental Results}

\label{app:extended-results}

We analyze the task-level outcomes for the same 1,375 tasks and ten configurations used in the main results. Whole-task success requires every included criterion to pass. Generation failures receive zero credit, including the task for which FLUX produced no image. Appendix~\ref{app:reproduction} describes the evaluation and aggregation procedures.

\subsection{B-VIS-Balanced and Identity-Level Results}

\paragraph{B-VIS-balanced domain results.}

For each represented B-VIS, we compute the whole-task pass rate among tasks containing it, then average these rates equally within its domain. This extends the five-model balanced profiles in the main text to all ten configurations. Unrepresented types are not assigned zero.

\begin{leviSupplementTable}
\caption{B-VIS-balanced whole-task pass rates (\%). Domain columns use equal weight per represented B-VIS.}
\label{tab:supp-balanced}
\centering\small
\setlength{\tabcolsep}{4pt}
\renewcommand{\arraystretch}{1.18}
\begin{tabular}{lrrrrrr}
\toprule
Configuration & Sym. & Geom. & Topo. & Layout & Quant. & Proc. \\
\midrule
GPT Image 2.5 Sunburst & 55.70 & 55.58 & 63.73 & 64.81 & 39.52 & 43.89 \\
GPT Image 2.5 Flare & 48.03 & 57.57 & 57.80 & 61.35 & 38.76 & 36.11 \\
GPT Image 2 & 49.46 & 49.72 & 55.82 & 54.63 & 28.04 & 37.49 \\
Qwen Image 3 Pro & 36.24 & 31.75 & 41.29 & 47.90 & 18.07 & 26.92 \\
Seedream 5 Pro & 27.55 & 32.74 & 33.08 & 32.17 & 13.75 & 15.59 \\
Nano Banana 2 & 29.93 & 31.05 & 28.87 & 28.03 & 14.11 & 25.18 \\
Nano Banana Pro & 28.75 & 30.20 & 29.17 & 33.52 & 15.75 & 20.85 \\
Grok Image 2 & 25.58 & 25.21 & 28.47 & 29.85 & 15.36 & 18.34 \\
MAI Image 2.5 & 24.80 & 23.93 & 18.91 & 27.95 & 9.05 & 11.94 \\
FLUX.2 Max & 17.68 & 13.77 & 15.03 & 13.23 & 4.84 & 5.42 \\
\bottomrule
\end{tabular}
\end{leviSupplementTable}

Flare reaches 57.57\% on Geometric under B-VIS balancing, compared with 55.58\% for Sunburst. This reverses their task-weighted ordering. Both aggregations measure complete-task success; neither isolates the accuracy of an individual component.

\paragraph{Within-domain profiles.}

To examine variation within these domain averages, Figure~\ref{fig:vis-performance} groups tasks by individual B-VIS. The two most frequent types per domain are selected independently of model performance.

\begin{leviSupplementFigure}

\centering

\includegraphics[width=\textwidth]{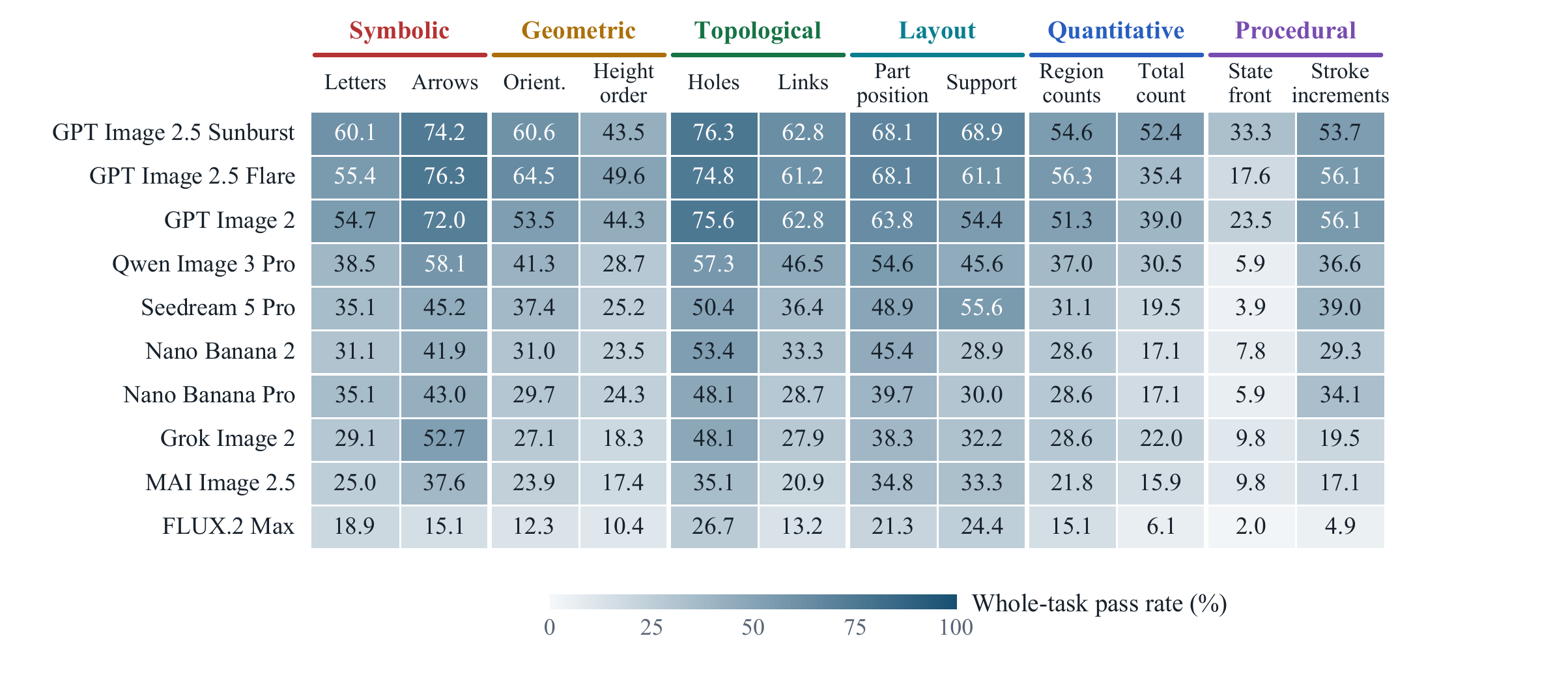}

\caption{Whole-task pass rates (\%) for the two most frequent B-VIS in each domain. Columns overlap when a task involves several types.}

\label{fig:vis-performance}

\end{leviSupplementFigure}

For Sunburst, orientation and height-order groups reach 60.65\% and 43.48\%, respectively. Flare leads Sunburst on partition counts (56.30\% versus 54.62\%) but trails on total count (35.37\% versus 52.44\%). Domain averages therefore conceal differences between identity types.

\subsection{Performance by Task Composition}

Table~\ref{tab:supp-composition} reports all configurations by the number of distinct B-VIS types and by domain involvement. These groups differ in task content as well as composition, so their pass-rate differences do not isolate the effect of composition.

\begin{leviSupplementTable}
\caption{Whole-task pass rates (\%) by task composition. The number of types refers to distinct B-VIS, not object instances.}
\label{tab:supp-composition}
\centering\small
\setlength{\tabcolsep}{4pt}
\renewcommand{\arraystretch}{1.18}
\begin{tabular}{lrrrrr}
\toprule
Configuration & \shortstack{1 type\\$n=143$} & \shortstack{2 types\\$n=421$} & \shortstack{3+ types\\$n=811$} & \shortstack{Single domain\\$n=367$} & \shortstack{Cross-domain\\$n=1008$} \\
\midrule
GPT Image 2.5 Sunburst & 62.24 & 56.06 & 59.43 & 55.31 & 59.92 \\
GPT Image 2.5 Flare & 52.45 & 52.97 & 57.71 & 50.95 & 57.44 \\
GPT Image 2 & 49.65 & 50.59 & 55.24 & 50.41 & 54.27 \\
Qwen Image 3 Pro & 42.66 & 37.53 & 40.32 & 37.60 & 40.48 \\
Seedream 5 Pro & 28.67 & 31.83 & 35.51 & 29.97 & 35.02 \\
Nano Banana 2 & 32.87 & 33.97 & 31.57 & 31.34 & 32.84 \\
Nano Banana Pro & 29.37 & 32.30 & 30.46 & 25.89 & 32.74 \\
Grok Image 2 & 23.78 & 25.18 & 30.70 & 26.16 & 29.07 \\
MAI Image 2.5 & 20.28 & 19.95 & 25.89 & 20.44 & 24.60 \\
FLUX.2 Max & 12.59 & 14.25 & 16.28 & 13.08 & 16.07 \\
\bottomrule
\end{tabular}
\end{leviSupplementTable}

\subsection{Shared Successes and Complementarity}

Table~\ref{tab:supp-difficulty} groups tasks by the number of configurations receiving a PASS. Each count summarizes the evaluated outputs from ten configurations; it does not estimate a model's success under repeated sampling.

\begin{leviSupplementTable}
\caption{Distribution of the number of passing configurations per task, out of ten.}
\label{tab:supp-difficulty}
\centering\small
\setlength{\tabcolsep}{4pt}
\renewcommand{\arraystretch}{1.18}
\begin{tabular}{lrr}
\toprule
Passing configurations & Tasks & Share (\%) \\
\midrule
0 & 243 & 17.67 \\
1 & 170 & 12.36 \\
2 & 132 & 9.60 \\
3 & 149 & 10.84 \\
4 & 142 & 10.33 \\
5 & 129 & 9.38 \\
6 & 145 & 10.55 \\
7 & 112 & 8.15 \\
8 & 60 & 4.36 \\
9 & 51 & 3.71 \\
10 & 42 & 3.05 \\
\bottomrule
\end{tabular}
\end{leviSupplementTable}

No configuration receives a PASS on 243 tasks (17.67\%); all ten receive a PASS on 42 (3.05\%). At least one receives a PASS on 1132 tasks. This is the combined coverage of the evaluated outputs; achieving it would require selecting a successful configuration separately for each task.

\begin{leviSupplementTable}
\caption{Paired task outcomes for the three highest-scoring configurations. A and B refer to the left and right names; every row sums to 1,375.}
\label{tab:supp-pairs}
\centering\small
\setlength{\tabcolsep}{4pt}
\renewcommand{\arraystretch}{1.18}
\begin{tabular}{lrrrr}
\toprule
A / B & Both pass & A only & B only & Neither \\
\midrule
Sunburst / Flare & 611 & 196 & 155 & 413 \\
Sunburst / GPT Image 2 & 602 & 205 & 130 & 438 \\
Flare / GPT Image 2 & 572 & 194 & 160 & 449 \\
\bottomrule
\end{tabular}
\end{leviSupplementTable}

Sunburst's net advantage over Flare is 41 tasks, comprising 196 Sunburst-only successes and 155 Flare-only successes. The aggregate ordering therefore does not imply that one model's successful tasks contain the other's.

\section{Evaluation and Reproducibility}

\label{app:reproduction}

Each configuration is evaluated on 1,375 tasks. Generation uses the complete text prompt without a vector reference. GPT-6-Sol with medium reasoning effort assesses each image against its task-specific criteria, as described in Section~\ref{sec:eval-setup}.

The challenge set was selected after generation based on redundancy, ambiguity, domain balance, and observed ease. The reported rates summarize automatic judgments on this selected set. They do not measure agreement with human judgments or variability across repeated generations.

For aggregation, assign one to PASS and zero otherwise, including generation failures. Average over all 1,375 tasks for the overall rate. Domain and B-VIS rates use the corresponding subsets defined by the task annotations; B-VIS-balanced rates give equal weight to each represented type. Reproduction requires the task prompt, generated image, model configuration, evaluation criteria, and judgment for each task. Image hashes in the supplementary source package identify the outputs used in the figures.

\end{document}